# Inferring Clinically Relevant Molecular Subtypes of Pancreatic Cancer from Routine Histopathology Using Deep Learning

**Abdul Rehman Akbar**[1, †, *], **Alejandro Levya**[1, †], Ashwini Esnakula[1], Elshad Hasanov[2], Anne Noonan[2], Lingbin Meng[2], Susan Tsai[7], Vaibhav Sahai[4, 5, 6], Midhun Malla[8], Sarbajit Mukherjee[9], Upender Manne[3], Anil Parwani[1], Wei Chen[1], Ashish Manne[2, ‡], Muhammad Khalid Khan Niazi[1, ‡]

[1] Department of Pathology, College of Medicine, The Ohio State University Wexner Medical Center, Columbus, OH, USA

[2] Department of Internal Medicine, Division of Medical Oncology, The Ohio State University Wexner Medical Center, Columbus, OH, USA

[3] O'Neal Comprehensive Cancer Center, The University of Alabama at Birmingham, Birmingham, AL, USA

[4] Department of Internal Medicine, Division of Hematology and Oncology, University of Michigan, Ann Arbor, MI

[5] Rogel Cancer Center, University of Michigan, Ann Arbor, MI

[6] Rogel and Blondy Center for Pancreatic Cancer, University of Michigan, Ann Arbor, MI

[7] Department of Surgery, The Ohio State University Wexner Medical Center and James Comprehensive Cancer Center, Columbus, OH, USA

[8] Division of Hematology and Oncology, University of Alabama at Birmingham, Birmingham, AL 35233, USA

[9] Miami Cancer Institute, Baptist Health South Florida, Miami, FL, USA

[†] The authors contributed equally to this work. The names are mentioned in alphabetical order.

‡ These authors jointly supervised this work.

[*] Correspondence: Abdul.Akbar@osumc.edu (A.R.A.)

## Abstract

**Background and aims:** Molecular subtyping of pancreatic ductal adenocarcinoma (PDAC) into basal-like and classical subtypes has established prognostic and predictive value. However, its adoption in routine clinical practice remains limited due to the cost, turnaround time, and tissue requirements of RNA sequencing. We introduce PanSubNet (PANcreatic SUBtyping NETwork), an interpretable deep learning framework that predicts therapy-relevant molecular subtypes directly from standard hematoxylin and eosin (H&E)–stained whole-slide images, enabling rapid and scalable molecular stratification.

**Methods:** PanSubNet was developed using data from 987 patients across two multi-institutional cohorts (PANCAN, n = 778; TCGA, n = 209) with paired histology and RNA sequencing data. Ground-truth labels were derived using the validated Moffitt 50-gene signature refined by GATA6 expression. The model employs a dual-scale architecture that integrates cellular morphology and tissue architecture through attention-based representation learning, enabling interpretable multi-scale feature extraction.

**Results:** On internal validation within PANCAN using five-fold cross-validation, PanSubNet achieved a mean area under the receiver operating characteristic curve (AUC) of 90.3% in high-confidence cases, with balanced sensitivity and specificity. External validation on the independent TCGA cohort without fine-tuning demonstrated robust generalizability (AUC 84.0%). Importantly, PanSubNet maintained meaningful predictive performance when applied to the entire PANCAN cohort, capturing the full biological continuum of the Moffitt subtype spectrum, including classical, basal-like, intermediate classical, and intermediate basal-like states. This capability enables clinically relevant stratification across both well-defined and transcriptionally ambiguous tumors.

**Conclusions:** By enabling rapid, cost-effective molecular subtyping from routine H&E slides while capturing the full continuum of PDAC transcriptional states, PanSubNet provides a clinically deployable and interpretable framework for molecular stratification, and supports integration into digital pathology workflows, advancing precision oncology for PDAC.

## Introduction

Pancreatic ductal adenocarcinoma (PDAC) is among the most lethal human malignancies. The overall 5-year survival for all-stage PDAC is approximately 11%, rising only to 42% for localized tumors and plummeting to 3% for metastatic tumors [1]. These outcomes stand in stark contrast to survival rates exceeding 90% for breast and prostate cancers, and approximately 65% for colorectal cancers. Despite this disproportionate mortality, systemic therapy selection in PDAC remains predominantly empiric. Across disease stages, treatment decisions are typically limited to two first-line regimens, (modified) FOLFIRINOX (irinotecan, oxaliplatin, and 5-fluorouracil) and gemcitabine/nab-paclitaxel (Gem-NP), with selection driven primarily by the patient fitness (performance status and age) rather than tumor biology [2-5]. This reflects both the absence of definitive comparative trials in advanced disease and the lack of clear superiority of either regimen in randomized studies of early-stage disease. Therapeutic progress over the past decade has been modest. Although the NAPOLI-3 trial showed improved outcomes with NALIRIFOX (nanoliposomal irinotecan, oxaliplatin, and 5-fluorouracil) compared to Gem-NP, survival metrics were comparable to those of FOLFIRINOX reported in the PRODIGE trial, underscoring the incremental rather than transformative gains achieved by irinotecan modification [6].

Routinely used biomarkers, including carbohydrate antigen 19-9 (CA 19-9), tissue mutation profiling, and cell-free DNA testing (liquid biopsy), provide limited prognostic insight and have little influence on treatment selection. Platinum-containing (cisplatin or oxaliplatin) regimens are preferentially considered for patients harboring select germline DNA damage repair alterations (e.g., *BRCA1/2, PALB2, ATM*, and *RAD50*); however, such alterations are rare, occurring in fewer than 10% of PDACs and up to 17% of familial cases [7]. Consequently, the vast majority of patients lack actionable biologic stratification at treatment initiation.

The clinical consequences of empiric therapy selection are most pronounced in settings with narrow therapeutic latitude. In potentially resectable tumors, ineffective initial treatment may permit progression to unresectable or locally advanced states, while in metastatic disease, early toxicity or rapid clinical decline may preclude subsequent lines of therapy. These constraints emphasize the need for earlier and more precise biologic stratification to inform treatment selection in PDAC.

Multiple molecular classification strategies, including mutational, proteomic, and transcriptomic signatures, have been proposed but remain largely confined to research [8-10]. Among them, transcriptomic subtyping into classical and basal-like phenotypes proposed by Moffitt et al., has demonstrated consistent prognostic ( clinical outcomes) and predictive (treatment response) relevance across cohorts [11]. The Purity Independent Subtyping of Tumors (PurIST) classifier developed by Rashid et al. provides a simplified, gene-pair-based implementation of classical and basal-like subtyping; however, its classifications are not fully concordant with the original Moffitt transcriptomic framework [12]. Classical PDAC is associated with superior outcomes and greater sensitivity to FOLFIRINOX, while basal-like PDAC portends poor prognosis and limited benefit from this regimen [11-22]. In a therapeutic landscape lacking biologically guided treatment algorithms, transcriptomic subtyping could, in principle, support treatment selection and prognostication. In practice, widespread clinical adoption of transcriptomic subtyping remains constrained by cost, turnaround times, limited global availability of RNA sequencing, and frequent tissue inadequacy, particularly in small diagnostic biopsies. Collectively, these constraints underscored the need for alternative strategies that can derive biologically meaningful feedback subtypes from routinely available clinical material.

Deep learning applied to routine hematoxylin and eosin (H&E)-stained whole-slide images (WSIs) offers a scalable digital pathology approach to inferring molecular insights. H&E slides are ubiquitously available from standard diagnostic workflows, and various deep learning-based studies have demonstrated a remarkable ability to infer molecular features, including gene expression signatures and mutations, directly from histology in several cancers [23-32]. By leveraging existing digital pathology infrastructure, such approaches enable rapid and cost-efficient molecular stratification compatible with routine clinical workflows, supporting broader clinical implementation in PDAC management.

Building on this foundation, we present PanSubNet (PANcreatic SUBtyping NETwork), a WSI–based deep learning framework for PDAC molecular subtyping. PanSubNet utilizes 1,055 patients with H&E WSIs and RNA-seq–derived labels using Moffitt 50-gene signature–based basal/classical scores refined by GATA6 expression as ground truth [33, 34]. The model operates at both cellular and tissue scales, integrating cell-level features with architectural context to generate slide-level subtype predictions (see Figure 1). By demonstrating that biologically meaningful PDAC subtype information, traditionally derived from transcriptomic profiling, can be inferred directly from routine histopathology, PanSubNet establishes WSIs as a scalable framework for biological stratification. In this study, PDAC molecular subtyping serves as a clinically relevant use case illustrating how WSI-based models extract tumor intrinsic biology from standard diagnostic materials, complementing existing molecular assays and remaining compatible with existing clinical workflows.

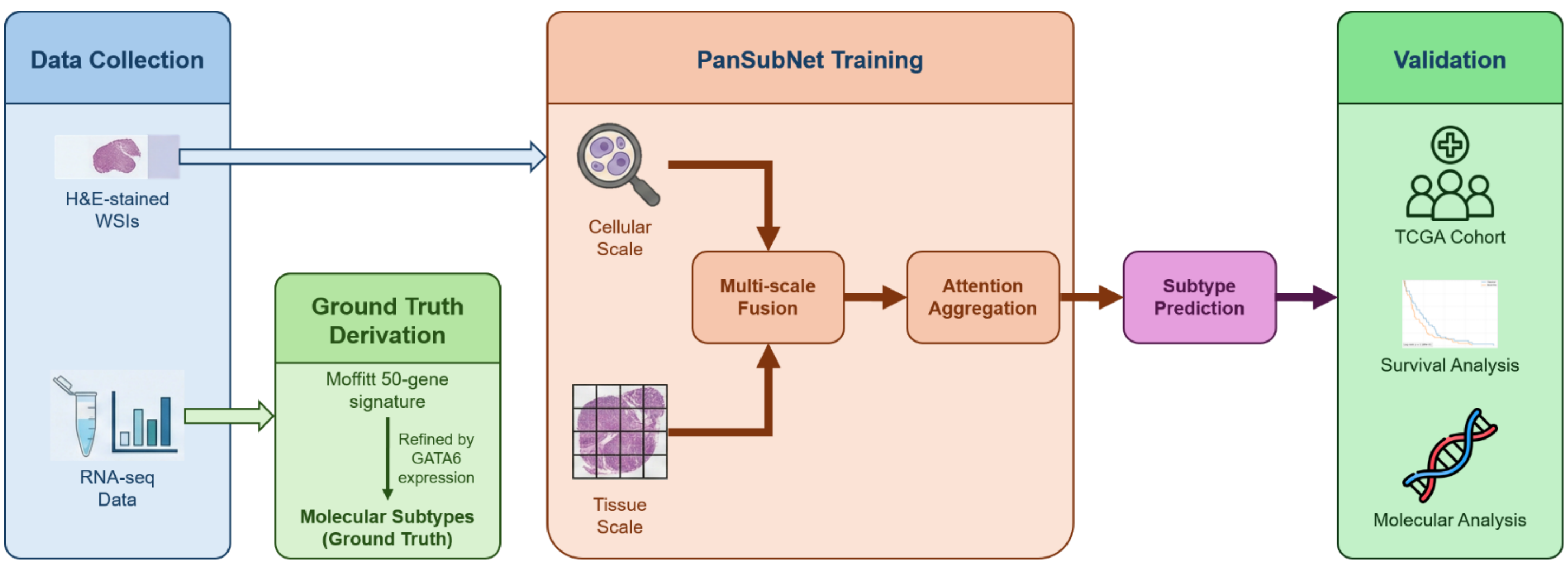

***Figure 1. Schematic overview of the PanSubNet study workflow.*** *The study methodology integrates paired H&E whole-slide images (WSIs) with transcriptomic data. Ground truth molecular subtypes are established using the Moffitt 50-gene signature, refined by GATA6 expression levels. The PanSubNet deep learning architecture employs a dual-scale approach to analyze the H&E slides: (1) a cellular scale utilizing CellViT++ to extract cell-level features and spatial context, and (2) a tissue scale utilizing UNI2-h to encode global patch-level features. These multi-scale representations are synthesized via a fusion and attention mechanism to generate a slide-level subtype prediction. The clinical and biological relevance of the model is validated through Kaplan-Meier survival analysis, assessment of DNA damage repair (DDR) gene associations, and functional pathway enrichment analysis.*

Recent advances in computational biology have demonstrated the feasibility of molecular features from WSIs using deep learning models [32, 35]. Prior approaches have primarily focused on learning associations between patch-level or slide-level image features molecular levels, establishing proof of concept for histology-based molecular inference across multiple cancer types. However, many existing methods treat molecular prediction as a hard classification problem and rely predominantly on regional correlations, with limited explicit modelling of cellular morphology, spatial organization, or biological uncertainty. Here, we address these considerations by developing a framework that integrates multi-scale histologic context for molecular inference from routine WSIs.

## Results

### Construction of high-confidence molecular labels across PANCAN and TCGA cohorts

We first derived transcriptomic ground-truth labels for molecular subtyping using Moffitt 50-gene basal-like and classical expression using techniques derived from previous literature [36, 37]. Single-sample gene-set enrichment analysis (ssGSEA) was used to compute subtype enrichment, and samples with |z-score| > 1 were designated as high-confidence basal-like or classical subtypes [38]. Intermediate cases were retained for downstream exploration but excluded from supervised model training.

Across the Pancreatic Cancer Action Network (PANCAN) cohort (n = 778), molecular data were available for 506 cases, including 171 high-confidence labels (95 basal-like and 76 classical) [39]. The 171 high-confidence samples were stratified based on clinical disease status, comprising 63 metastatic cases and 108 surgically resected cases. Within The Cancer Genome Atlas (TCGA) cohort (n = 209), molecular data were available for 183 cases, with 62 high-confidence classifications (15 basal-like and 47 classical) (see **Supplementary Figure S1**). These high-confidence subsets formed the basis for model training, internal validation, and external testing. The distribution of high-confidence subtypes was consistent with previously reported proportions for PDAC transcriptomic classes and reflects the predominance of the classical lineage in bulk RNA sequencing datasets [40].

**PanSubNet accurately predicts basal-like and classical subtypes in internal validation on PANCAN**

PanSubNet was trained exclusively on the high-confidence PANCAN cases and evaluated using a five-fold cross-validation framework. The model demonstrated strong discriminatory ability, achieving a mean AUC of 90.302 ± 5.331 and an accuracy of 87.020 ± 10.707 across folds (see **Figure 2**, **Supplementary Table S1**). Balanced accuracy (87.181 ± 10.862) indicated that performance was stable across both subtypes, with no substantial bias despite the modest class imbalance. Sensitivity and specificity were similarly well-aligned (86.795 ± 13.551 and 87.567 ± 8.331, respectively), suggesting that PanSubNet reliably distinguishes basal-like from classical tumors even within a biologically heterogeneous cohort.

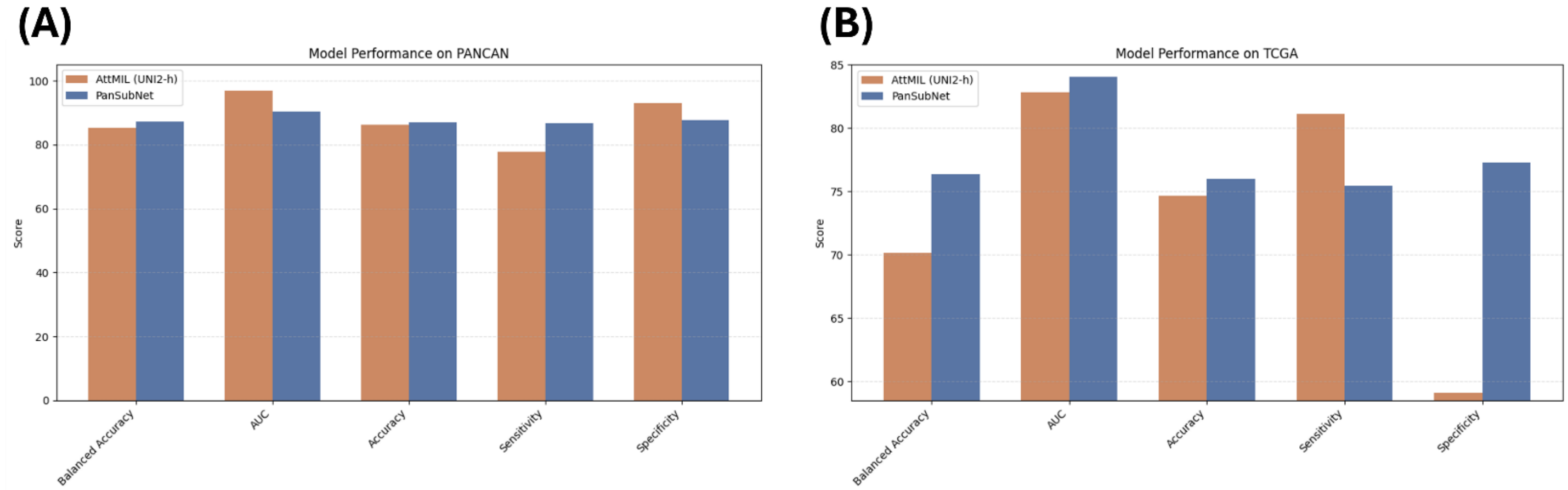

***Figure 2. Performance comparison of PanSubNet and AttMIL across internal and external cohorts.*** *Comparison of PanSubNet and AttMIL (UNI2-h backbone) performance on the PANCAN and TCGA cohorts. Metrics shown include area under the receiver operating characteristic curve (AUC), accuracy, balanced accuracy, sensitivity, and specificity.*

Across 171 samples, the model achieved an overall concordance (accuracy) of 0.889, with 19 misclassifications. Error rates were comparable between subtypes, with 9/95 errors in basal tumors (9.5%) and 10/76 errors in classical tumors (13.2%), indicating no strong class-specific bias. Correct predictions exhibited significantly higher decision margins than incorrect predictions (median margin 0.462 vs. 0.384; Mann–Whitney U test, $p = 0.014$), suggesting that misclassifications are associated with reduced model confidence. Notably, 6 of 19 errors (31.6%) occurred with high predicted confidence ($p \leq 0.10$ or $\geq 0.90$), consistent with a subset of confidently misclassified cases rather than uniformly ambiguous predictions. Concordance of RNA sequencing and PanSubNet is discussed in **Supplementary Table S3**.

**PanSubNet demonstrates translationally relevant performance across the complete Moffitt subtype spectrum**

For clinical deployment, a subtype prediction model must operate across the full biological continuum observed in practice, rather than only on transcriptomically well-defined tumors. In Moffitt classification system, cases often fall along a spectrum spanning classical, basal-like, intermediate classical, and intermediate basal-like phenotypes. To support real-world clinical decision-making, PanSubNet must therefore generalize beyond high-confidence cases and differentiate tumors across this complete spectrum.

To assess this capability, we conducted an exploratory analysis in which the trained model, learned exclusively from high-confidence PANCAN cases, was applied at inference time to the entire PANCAN cohort, including samples with low-confidence subtype scores. As anticipated, the inclusion of biologically ambiguous tumors resulted in a measurable reduction in predictive performance. Across five-fold cross-validation, PanSubNet achieved an AUC of 71.307 ± 6.370, accuracy of 68.580 ± 4.295, balanced accuracy of 68.422 ± 4.369, sensitivity of 69.952 ± 5.779, and specificity of 66.893 ± 7.056 (see **Supplementary Table S1**).

Furthermore, to determine whether PanSubNet representations capture signals related to subtype confidence, we fine-tuned a small set of additional layers on top of the learned PanSubNet embeddings to classify cases as high-confidence versus low-confidence according to transcriptomic subtype scores. This auxiliary model demonstrated consistent discriminative ability, achieving an AUC of 74.9 ± 5.30, accuracy of 74.2 ± 4.75, balanced accuracy of 73.9 ± 5.00, sensitivity of 72.5 ± 8.67, and specificity of 75.3 ± 6.29, indicating that PanSubNet representations capture meaningful biological signals associated with subtype certainty (see **Supplementary Table S1**).

**PanSubNet demonstrates strong generalizability in external validation on TCGA**

To assess cross-cohort generalizability, PanSubNet trained exclusively on PANCAN WSIs was evaluated on the high-confidence TCGA PDAC cohort without any fine-tuning. Despite differences in institutional origin, sequencing platforms, slide preparation, and staining variability, the model maintained high performance, achieving an AUC of 84.048 and accuracy of 76.000 (see **Figure 2**, **Supplementary Table S1**). Balanced accuracy (76.372) demonstrated that classification remained stable across subtypes, and sensitivity and specificity (75.472 and 77.273, respectively) confirmed robust external performance.

**Benchmarking against baseline reveals superior cross-cohort stability of PanSubNet**

We compared PanSubNet against the attention-based multiple instance learning model AttMIL  with UNI2-h, a widely used WSI encoder [41, 42]. On internal evaluation using PANCAN, AttMIL achieved a similar balanced accuracy (85.336 ± 4.718) and higher AUC (96.812 ± 1.026), but at the expense of a substantial imbalance between sensitivity and specificity. This imbalance reflects a bias toward predicting the classical subtype, resulting in under-calling basal-like cases. In external validation on TCGA, AttMIL exhibited a marked reduction in performance, with AUC decreasing to 82.847, accuracy to 74.667, and specificity to 59.091. In contrast, PanSubNet preserved high discriminative performance on TCGA, with stable sensitivity and specificity across cohorts (see **Figure 2, Supplementary Table S2).**

**Molecular subtypes stratify patients by overall survival in the PANCAN cohort**

We evaluated the association between molecular subtypes and overall survival (OS) in the PANCAN cohort using Kaplan–Meier analysis under multiple stratification settings (see Figure 3). Analyses were performed separately for patients who developed metastatic disease and for the full cohort, including both metastatic and resected patients. Subtype assignments were derived from both RNA-seq–based molecular labels and PanSubNet model predictions to enable side-by-side comparison.

Among metastatic patients, OS stratification was first evaluated using high-confidence RNA-seq–derived molecular subtype labels. In this high-confidence subset, Kaplan–Meier analysis showed separation between classical and basal-like tumors; the difference had a trend towards statistical significance (p=0.08, see **Figure 3a**). When the same metastatic subset was stratified using PanSubNet-predicted subtypes, overall survival differed significantly between groups (see **Figure 3b**). Notably, several cases labeled as classical by RNA-seq were predicted as basal-like by the model, and these misclassified cases experienced early death events. Consequently, the median OS of the classical group differed between labeling strategies, with a median OS of 22.0 months using RNA-seq–derived labels compared to 24.7 months using model-predicted labels. This preservation of survival stratification using histology-only inference indicates that PanSubNet captures clinically relevant tumor biology within routine WSIs.

We next assessed OS in all high-confidence cases, including both metastatic and resected patients. In this setting, RNA-seq–derived molecular subtypes demonstrated modest separation in OS that did not consistently reach statistical significance (see **Figure 3c**). Stratification of the same cohort using PanSubNet-predicted subtypes likewise showed limited separation between groups, with no statistically significant differences observed (see **Figure 3d**).

To examine the impact of subtype confidence on survival associations, we additionally repeated OS analyses in the full PANCAN cohort, including both high- and low-confidence cases (see **Supplementary Figure S2**). Across these analyses, inclusion of low-confidence cases was associated with reduced separation between survival curves compared to analyses restricted to high-confidence molecular labels. The confidence-dependent improvement in the prognostic signal suggests that internal estimates meaningfully reflect biological fidelity rather than model noise.

In summary, survival differences between RNA-sequencing–defined subtypes were more apparent in the metastatic subgroup than in the full cohort, which includes resected early-stage cases, highlighting the confounding influence of disease stage and surgical resection on overall survival. Notably, PanSubNet preserved robust prognostic stratification in metastatic patients and demonstrated stronger survival separation than RNA-seq–based subtype labels in this setting.

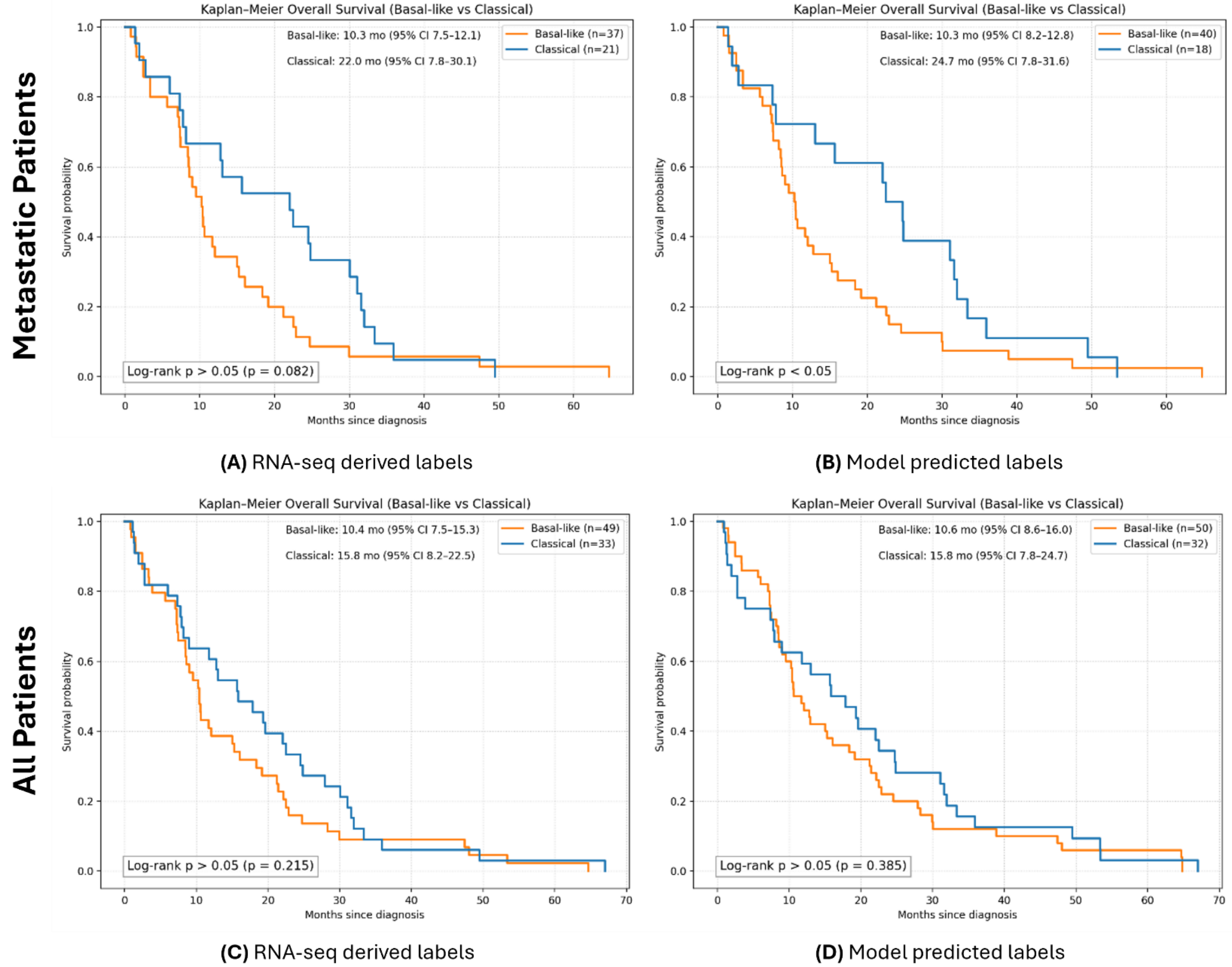

***Figure 3. Overall survival stratified by molecular subtype in the PANCAN cohort (high-confidence cases).*** *Kaplan–Meier curves showing overall survival (OS) stratified by Basal-like and Classical subtypes across different patient subsets and labeling strategies.* ***(A)*** *OS among patients who developed metastatic disease, stratified using RNA-seq–derived molecular subtype labels.* ***(B)*** *OS among the same metastatic patients, stratified using PanSubNet-predicted molecular subtypes.* ***(C)*** *OS in all high confidence cases (metastatic and resected), stratified using RNA-seq–derived molecular subtype labels.* ***(D)*** *OS in all high confidence cases, stratified using PanSubNet-predicted molecular subtypes. Median survival times with 95% confidence intervals are reported within each panel. Log-rank test results are indicated on the plots.*

## Discussion

Our findings demonstrate that the basal-like and classical subtypes of PDAC, originally defined through transcriptomic profiling, manifest as distinct and morphologically learnable histopathological patterns. This observation is consistent with fundamental principles of tumor biology: transcriptional programs shape cellular phenotypes, and these phenotypes leave measurable morphological signatures within tissue architecture that can be captured in routine H&E-stained sections [43]. The ability of PanSubNet to infer RNA-seq–defined molecular subtypes directly from histology supports the premise that transcriptional identity is encoded in tissue organization, cellular morphology, and microenvironmental context. This, to our knowledge, represents the first WSI-based PDAC subtyping according to the Moffitt classification system. Prior studies were based on PurIST classifier, which demonstrates a incomplete concordance with the Moffitt system [15, 32, 35, 44]. A foundational step of this study was establishing and validating the molecular ground truth used to train PanSubNet. Since PanSubNet is explicitly optimized to predict RNA-sequencing–derived basal-like versus classical subtype labels, all transcriptomic analyses were intentionally performed using RNA-seq–derived subtype assignments rather than model-predicted labels. This design avoids circular inference: using PanSubNet predictions to characterize transcriptomic programs would confound biological interpretation, as any observed differences would primarily reflect model optimization rather than independent molecular structure. Accordingly, RNA-seq analyses in this work serve exclusively to confirm that the supervised target labels correspond to coherent biological programs and clinically meaningful phenotypes, rather than to introduce a parallel molecular discovery effort. Detailed transcriptomic characterizations, including lineage markers, pathway enrichment, and subtype ambiguity analyses, are therefore presented in the **Supplementary Methods 1-6**, while the primary Results section focuses on histology-based prediction performance and clinical associations.

Within these RNA-seq–based validation analyses, GATA6 emerged as a robust lineage-associated marker, exhibiting elevated expression in classical tumors and reduced or absent expression in basal-like tumors, including a subset of samples with undetectable expression (**Supplementary Methods 3**). This distribution is consistent with prior literature and reinforces GATA6's role as a marker of pancreatic epithelial differentiation. Importantly, GATA6 is not proposed here as an alternative subtyping framework, but rather as a biological anchor that corroborates the transcriptional identity of the classical subtype and helps contextualize intermediate or ambiguous cases [16, 33]. Together with pathway-level analyses, these findings confirm that the RNA-seq–derived subtype labels used for model supervision represent stable, biologically grounded endpoints.

Ambiguity analyses further revealed that PDAC molecular subtypes exist along a continuous transcriptional spectrum, with basal-like and classical tumors occupying relatively stable endpoints and intermediate cases forming a transitional zone. This biological continuum presents an inherent challenge for supervised classification, as no sharp boundary separates subtypes. By training exclusively on high-confidence cases, where transcriptional identity is unambiguous, PanSubNet learns the clearest morphological correlates of the subtype extremes. The model's attention mechanisms then enable interpolation across this spectrum, capturing patterns consistent with partial differentiation or incomplete dedifferentiation. The strong external validation performance across institutions, staining protocols, and specimen types suggests that these learned patterns generalize beyond cohort-specific artifacts (see **Figures 4** and **5**).

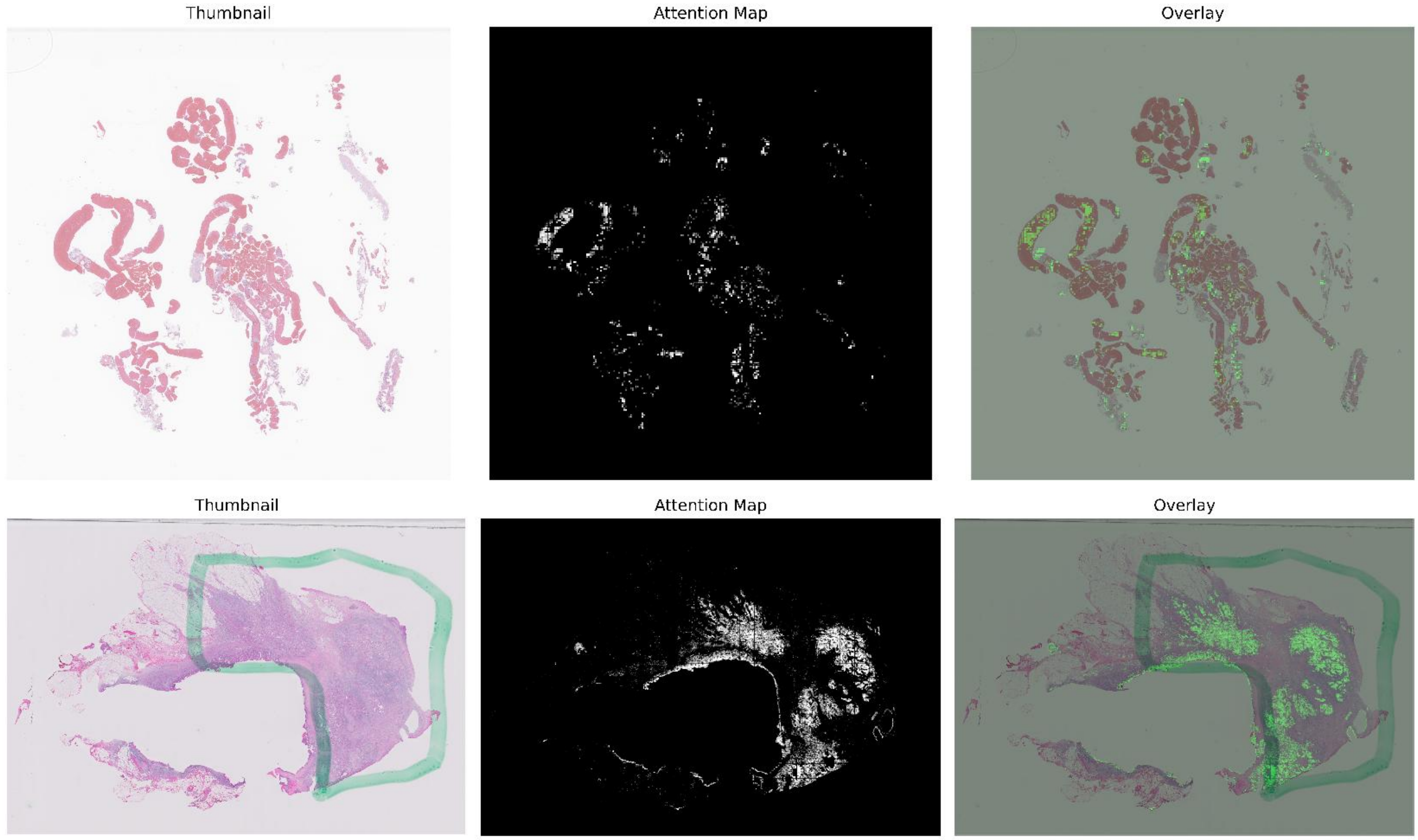

***Figure 4. PanSubNet attention maps highlight discriminative regions in classical subtype patients.*** *Attention maps from PanSubNet for fine needle biopsy and surgical resection specimens, from two randomly sampled, correctly predicted classical-subtype patients. The left panel shows a thumbnail of the original whole-slide image. The middle panel displays the attention mask, with values ranging from 0 (black) to 255 (white), where white indicates the highest attention. The right panel shows the attention map overlaid on the original*

*slide.*

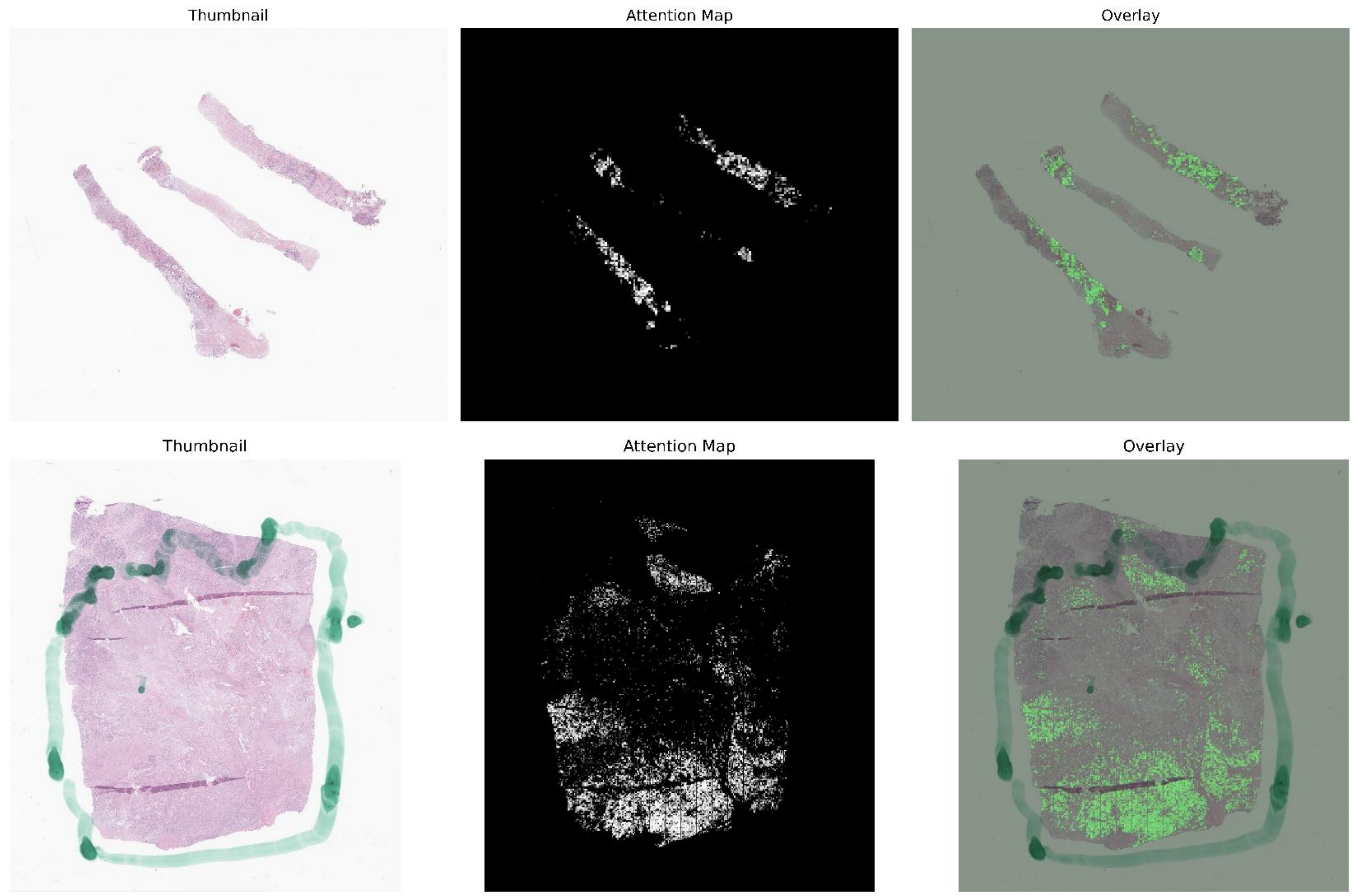

***Figure 5. PanSubNet attention maps identify basal-like–associated regions.*** *Attention maps from PanSubNet for fine needle biopsy and surgical resection specimens, from two randomly sampled, correctly predicted basal-like-subtype patients. The left panel shows a thumbnail of the original whole-slide image. The middle panel displays the attention mask, with values ranging from 0 (black) to 255 (white), where white indicates the highest attention. The right panel shows the attention map overlaid on the original slide.*

Consistent with prior studies, classically subtyped tumors by RNA-sequencing exhibited more favorable survival than basal-like tumors, with a trend towards significance in metastatic patients (p=0.08) [45]. Our analyses reveal that subtype differences are most pronounced in metastatic cases, where tumor biology exerts an influence outcome. PanSubNet subtype assignments maintained, and even improved survival stratification compared to RNA-seq labels. Some tumors labeled classical by RNA-seq but basal-like by PanSubNet, which had early deaths, suggest histomorphology at the whole-slide level reflects aggressive biology not fully captured by bulk transcriptomics. Including resected patients reduced survival differences across subtypes for both methods, highlighting the influence of surgery and disease stage on long-term survival outcomes. Transcriptomic analyses linking subtype status with DNA damage repair (DDR) gene expression provide a plausible biological rationale for these clinical differences and suggest that integrating subtype information with DDR-related biomarkers may ultimately support more refined therapeutic stratification (**Supplementary Methods 5**). However, such integrative approaches remain exploratory and were not evaluated directly in this study.

PanSubNet represents a step toward democratizing molecular stratification in PDAC. Unlike RNA sequencing, which requires specialized infrastructure, sufficient tissue yield, and prolonged turnaround, PanSubNet operates directly on H&E slides, the standard output of diagnostic pathology workflows worldwide. WSIs can be scanned, analyzed, and classified within hours, offering subtype information on a clinically relevant timescale. This rapid turnaround is particularly important in PDAC, where disease progression often necessitates urgent treatment decisions.

The model's robust external performance on TCGA (AUC 0.84) demonstrates resilience to inter-institutional variability, an essential prerequisite for clinical deployment. Notably, PanSubNet maintained balanced sensitivity and specificity across cohorts, avoiding the classical-subtype bias observed in the AttMIL baseline. This suggests that the multi-scale, cell-to-patch fusion architecture more effectively captures biologically meaningful features than patch-level aggregation alone. Misclassification rates were similar for basal-like and classical tumors, with errors mainly in ambiguous cases, not systematic bias. Some discordant cases likely show true histologic–transcriptomic divergence, possibly due to heterogeneity, sampling, or lineage plasticity. These findings support PanSubNet as a robust, biologically grounded WSI-based subtyping approach.

Several limitations warrant consideration. The ground-truth labels were derived using the Moffitt 50-gene signature with ssGSEA scoring. While well-validated, alternative PDAC subtyping frameworks such as PurIST exist, and the optimal classification schema remains debated [46-48]. Training PanSubNet using alternative label definitions may clarify whether the model learns generalizable morphological correlates of tumor biology or schema-specific patterns [49].Although PanSubNet was developed and evaluated in a cohort exceeding 1000 patients, PDAC is a biologically and histologically heterogeneous disease; therefore, further validation in larger, more diverse, multi-institutional datasets will be important for assessing generalizability.

We acknowledge that molecular subtyping alone has limited immediate impact on treatment selection in the current PDAC therapeutic landscape, where first-line options remain restricted. Nevertheless, subtype assignment provides meaningful prognostic context that may support anticipatory clinical planning. For basal-like tumors, this includes early recognition of aggressive biology and prioritization of clinical trial enrollment; for classical tumors, it may support toxicity-sparing strategies and sustained therapy continuation. In its current iteration, PanSubNet does not currently integrate clinical variables or immunohistochemical markers. Hybrid models combining histology-based inference with targeted IHC or clinical features may improve interpretability and risk stratification but are beyond the scope of this work. Future work will focus on prospective validation, integration with orthogonal biomarkers, and extension of histology-based molecular inference to additional tumor types where transcriptional subtyping informs therapy response [50]PanSubNet is intended to complement current clinical and pathological evaluations, and is not designed to provide treatment recommendations or function as a clinical decision-support tool.

In summary, our PanSubNet classifies PDACs into clinically useful subtypes directly from the routine H&E slides with biological and clinical fidelity. By lowering barriers to molecular stratification, it enables scalable integration of tumor biology into translational research and clinical workflows. clinical contexts where transcriptomic profiling is impractical, including small biopsies with insufficient RNA yield, community hospitals without access to sequencing platforms, rapid triage in metastatic disease, retrospective stratification of archival specimens for clinical trials, and intraoperative or cytology specimens.

## Materials and Methods

### Datasets and RNA-sequencing preprocessing

Our study comprised 987 patients across two cohorts: the PANCAN cohort (n = 778) and the TCGA-PAAD cohort (n = 209). All patients had diagnostic formalin-fixed, paraffin-embedded (FFPE) whole-slide images (WSIs) scanned at ×40 magnification. From this total, 689 patients (506 from PANCAN, 183 from TCGA) had paired H&E WSIs and bulk RNA sequencing data, which formed the dataset for downstream analysis.

For TCGA-PAAD, raw RNA-seq data were obtained through the Genomic Data Commons (GDC) manifest system. Expression values were downloaded as gene-level count matrices and converted to transcripts per million (TPM) using GENCODE gene models. Gene identifiers (ENSEMBL) were mapped to standard HUGO gene symbols using the corresponding GTF annotation [51, 52]. Expression values were extracted specifically for the 50-gene Moffitt classical/basal signature, and TPM values were z-scored across samples to normalize gene-level variation.

For the PANCAN cohort, raw FASTQ files were downloaded and processed following a consistent, uniform RNA-seq pipeline. Firstly, QC was performed on raw FASTQ files to assess GC content, sequence duplication, read saturation, and potential contamination. Secondly, transcript quantification was performed using Salmon (v1.10.2) in quasi-mapping mode with the appropriate Salmon index constructed from the same GENCODE reference used for TCGA [53]. Thirdly, transcript counts across all transcript isoforms were aggregated into gene-level abundances. Resulting TPM matrices were cross-checked to ensure complete coverage of all 50 Moffitt genes.

For both cohorts, single-sample gene-set enrichment analysis (ssGSEA) was used to compute classical and basal-like enrichment scores based on the Moffitt 50-gene signature. For each sample, genes were ranked by TPM; ssGSEA enrichment was computed separately for classical and basal gene sets; and a molecular subtype score was calculated as:

$$score = ssGSEA(classical) - ssGSEA(basal)$$

Positive values indicate classical-like expression; negative values indicate basal-like expression.

Subtype scores were then z-scored across all 792 patients to quantify the confidence of each classification.

We defined:

$$Subtype = \begin{cases} Classical, & z-score > +1 \\ Basal-like, & z-score < -1 \\ Intermediate, & -1 \leq z-sccore \leq +1 \end{cases}$$

This thresholding reproduces the expected ~65/35 classical/basal distribution reported in prior studies and yields robust subtype assignments suitable for supervised learning.

Across the full cohort, 233 patients met high-confidence criteria (110 basal-like, 123 classical), and 456 patients fell into the intermediate range. High-confidence cases were used for all supervised training, internal validation, and external testing. Intermediate cases were excluded from model training but retained as a biological reference group for other downstream analysis tasks.

GATA6 is a well-established marker of the classical lineage and distinguishes classical from basal tumors in multiple transcriptomic frameworks. To refine ambiguous intermediate cases, GATA6 TPM values were stratified into tertiles (first tertile ≈ ≤20 TPM; third tertile ≈ ≥60 TPM). Cases that fell between the two tertiles were labeled as ambiguous. In the combined cohort, no ambiguous cases remained after using GATA6 expression filtration. Samples in the lowest tertile were labeled basal-like, while the samples in the highest tertile were labeled classical. This refinement reflects clinical practice in which classical-like tumors with preserved differentiation (high GATA6) are considered biologically distinct from poorly differentiated, basal-like tumors.

After ssGSEA scoring, z-scoring, and GATA6 refinement, PANCAN (n = 506) contributed 171 high-confidence cases, out of which 95 were high-confidence basal-like, while 76 were high-confidence classical. On the other hand, TCGA (n = 183) contributed 62 high-confidence cases, out of which 15 were high-confidence basal-like, while 47 were high-confidence classical. These 233 high-confidence samples (171 PANCAN, 62 TCGA) served as the ground-truth for supervised model training and external testing. The resulting subtype proportions and classical/basal ratios were consistent with published PDAC transcriptomic datasets and maintain biological fidelity for model development.

**Overview of PanSubNet architecture**

We propose PanSubNet, a deep learning framework that implements the language of histopathology, as proposed in our recent study [23], by integrating information across multiple scales and explicitly modeling cellular and intercellular characteristics within their biological context. The architecture is composed of three core components that mirror natural language processing (NLP) architectures: (1) multi-scale feature extraction to capture both cellular "words" and tissue patch "sentences"; (2) cell-to-patch mapping and fusion to create contextual embeddings; and (3) multi-dimensional attention mechanisms to learn complex semantic relationships between tissue regions.

Our strategy implements a dual-scale approach, where cells at 40× magnification serve as fundamental "words" encoding fine-grained morphological and phenotypic information, and tissue patches at 20× magnification represent "sentences" providing architectural context and spatial organization. Each WSI was tiled into non-overlapping 256 x 256-pixel patches at 20× magnification, and for each patch a feature vector was extracted using the pre-trained foundation model UNI2-h [41], capturing broad morphological and tissue-architectural patterns. Concurrently, cell nuclei were segmented and classified from WSIs at 40× magnification using CellViT++, specifically the SAM-based model [54-56]. This model yields embeddings for each individual cell, encoding fine-grained morphological and phenotypic details across five major cell categories, along with centroid coordinates for each cell.

To implement contextual embedding principles, we established a spatial correspondence between cells and patches by mapping each cell to the patch containing its centroid. Within each patch, we aggregated variable-length cell embeddings using a spatially biased self-attention mechanism with a learnable CLS token [57]. To incorporate spatial information, the physical Euclidean distance between cell centroids was subtracted from the raw attention scores. This spatial bias encodes biological function, where cellular proximity defines semantic relationships and functional neighborhoods. The CLS token was extracted, containing aggregated information from all cells within the patch with greater influence from spatially proximal cells.

To integrate cellular information with tissue architecture, we employed an outer product operation between the patch embedding vector and the aggregated cell embedding vector (CLS token) for each patch. This resulted in a matrix that captures all pairwise multiplicative interactions between patch and cell features. This representation was flattened and projected back to 768 dimensions using a learnable projection matrix to obtain the final fused representation. This fusion mechanism enables rich cross-scale interactions that capture how cellular characteristics interact with tissue architecture.

The fused embeddings from a WSI form a "bag" of k instances, which were aggregated using 2D AttMIL for slide-level prediction, as proposed in our recent study [23].

### Model training and evaluation

PanSubNet was trained exclusively on the 171 high-confidence PANCAN cases using five-fold cross-validation. In each fold, the model was trained to classify basal-like versus classical subtypes using binary cross-entropy loss, optimized with AdamW optimizer (learning rate 5e-05, weight decay 1e-05). Models were trained for up to 100 epochs with early stopping based on validation AUC. The final reported metrics represent the mean and standard deviation across all five folds.

For external validation, the model having best validation performance among five-fold PANCAN validation sets, was evaluated on the 62 high-confidence TCGA cases without any fine-tuning or retraining. This zero-shot evaluation assessed true cross-institutional generalizability. Performance metrics included area under the receiver operating characteristic curve (AUC), accuracy, balanced accuracy, sensitivity (recall for basal-like), and specificity (recall for classical).

### Baseline comparison

We compared PanSubNet against AttMIL [42], a standard architecture using UNI2-h patch embeddings without cellular information. AttMIL aggregates patch-level features through a single attention layer followed by a classification head. This baseline was trained and evaluated using identical data splits and preprocessing to enable direct comparison.

### Survival analysis

Kaplan–Meier survival analysis was performed in the PANCAN cohort to evaluate associations between molecular subtypes and OS. Analyses were conducted under multiple stratification settings, including patients who developed metastatic disease and the full cohort comprising both metastatic and resected patients. Molecular subtype assignments were obtained using either RNA-seq–derived molecular labels or PanSubNet model–predicted labels, enabling direct side-by-side comparison of labeling strategies. OS was defined as the time interval from the date of initial diagnosis to the date of death. Patients with documented death events were included in the analysis, while patients who were alive but lost to follow-up were excluded. Survival times were measured in months. Survival curves were estimated using the Kaplan–Meier method, and differences between groups were assessed using the log-rank test. Median overall survival values and corresponding 95% confidence intervals were computed for each group and reported in the figures. To evaluate the impact of subtype confidence, survival analyses were performed both in high-confidence molecular subtype cases and in the full PANCAN cohort including high- and low-confidence cases.

### Gene ontology and pathway enrichment analysis

Gene Ontology (GO) Biological Process and KEGG pathway enrichment analyses were performed on the top 25 most discriminative genes for each subtype (classical and basal-like) using the clusterProfiler R package. Enrichment p-values were calculated using hypergeometric tests with Benjamini–Hochberg correction for multiple testing. Significantly enriched terms (adjusted $p < 0.05$) were visualized as dot plots with −log10(p-value) on the x-axis. Gene-to-term networks were constructed using the enrichplot package to visualize the relationships between genes and their associated GO terms or KEGG pathways. These analyses can be found in Supplementary Materials.

### DNA damage repair gene expression analysis

Expression of key DDR genes (BRCA1, BRCA2, PALB2, RAD51, ATM, CHEK1) accounting for mutations, was extracted from the TPM matrices for all samples. A composite DDR score was calculated by z-scoring each gene across all samples and summing the z-scores. DDR gene expression was compared between subtypes using violin plots and heatmaps. Correlation analyses assessed the relationship between DDR score and subtype confidence ($\Delta Z$) and prediction ambiguity. These analyses can be found in the Supplementary Materials.

**Code and data availability**

The underlying code for this study is available in AI4Path PanSubNet repository and can be accessed via this link . WSIs and RNA-seq data for TCGA-PAAD are publicly available through the Genomic Data Commons (https://portal.gdc.cancer.gov/). PANCAN data access is subject to institutional data use agreements. Processed data and model weights will be made available upon reasonable request to the corresponding author.

**Author Contributions**

A.R.A. and A.L. contributed equally to this work. A.R.A. designed and implemented the PanSubNet architecture, performed computational analyses, and wrote the manuscript. A.L. performed RNA sequencing preprocessing, molecular characterization, gene ontology analyses, and contributed to manuscript writing. A.E. provided pathology expertise, reviewed the attention maps of the model, and contributed to manuscript editing and revision. A.P. and E.H. provided oversight and resources for the study and contributed to manuscript editing and revision. W.C. supervised the clinical side of the study, participated in model training to ensure clinical relevance, and contributed to manuscript editing and revision. A.M. contributed to study conception, study design, and analytical strategy, provided expert guidance on pancreatic cancer biology, molecular subtyping, and clinical relevance, and contributed to manuscript writing, editing, and revision. M.K.K.N. conceptualized, designed, and validated the study, supervised the research, provided funding, and edited and revised the manuscript.

**Competing Interests**

S.M. received research funding from the National Comprehensive Cancer Network and Ipsen Biopharmaceuticals/North American Neuroendocrine Tumor Society, which were paid to the institute and outside the scope of this work. SM received consulting fees from Merck, Eisai, BMS, Boehringer Ingelheim, Gilead, Natera, Incyte and BeiGene/BeOne Ltd. All other authors declare no competing interests.

**Acknowledgments**

We thank the patients who contributed samples to the PANCAN and TCGA cohorts. We acknowledge the Genomic Data Commons for providing public access to TCGA data. We also gratefully acknowledge the Ohio Supercomputer Center for providing high-performance computing resources as part of its contract with The Ohio State University College of Medicine. We also thank the Department of Pathology and the Comprehensive Cancer Center at The Ohio State University for their support.

**Funding**

The project described was supported in part by R01 CA276301 (PIs: Niazi and Chen) from the National Cancer Institute, Pelotonia under IRP CC13702 (PIs: Niazi, Vilgelm, and Roy), The Ohio State University Department of Pathology and Comprehensive Cancer Center. The content is solely the responsibility of the authors and does not necessarily represent the official views of the National Cancer Institute or National Institutes of Health or The Ohio State University.

**Ethics Approval and Consent to Participate**

This study involved secondary analysis of retrospective, fully de-identified clinical, molecular, and histopathology data obtained from existing institutional and public repositories. In accordance with applicable regulations, the use of de-identified data does not constitute human subjects' research. Therefore, Institutional Review Board (IRB) approval was not required, and informed consent to participate was waived.

**Supplementary Materials**

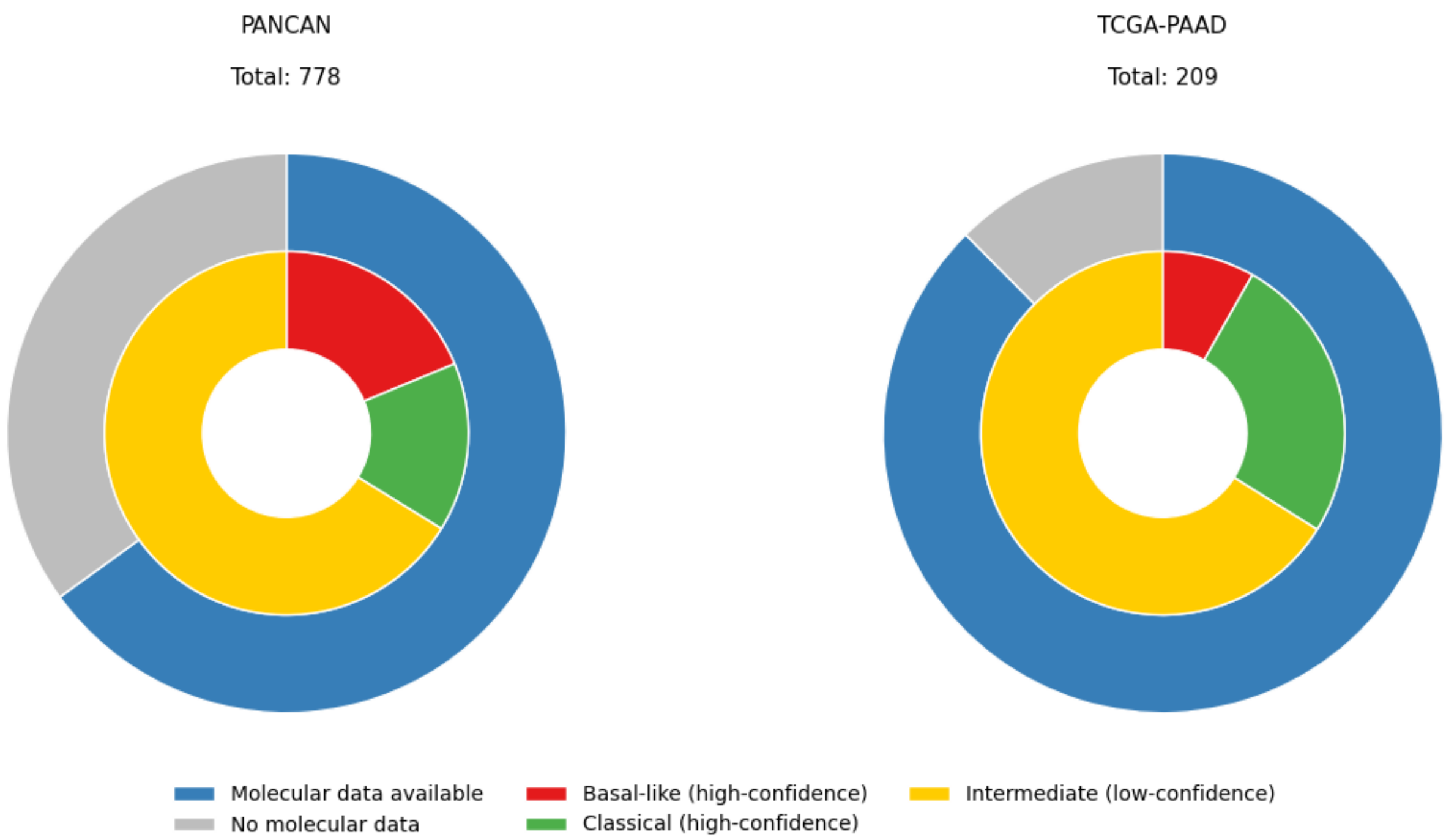

***Figure S1:*** *Nested donut charts illustrating cohort composition and molecular subtype distribution. The outer ring represents the proportion of patients with available molecular data (blue) versus those without (gray) for each cohort (PANCAN and TCGA-PAAD). The inner ring shows the breakdown of molecular cases into high-confidence subtypes: basal-like (red) and classical (green) and intermediate/low-confidence cases (yellow).*

***Table S1. Performance of PanSubNet for molecular subtype prediction across cohorts.***

| Dataset | AUC | Accuracy | Bal. Acc. | Sensitivity | Specificity |
|---|---|---|---|---|---|
| **PANCAN (high confidence)** | 90.302 ± 5.331 | 87.020 ± 10.707 | 87.181 ± 10.862 | 86.795 ± 13.551 | 87.567 ± 8.331 |
| **PANCAN (entire cohort)** | 71.307 ± 6.370 | 68.580 ± 4.295 | 68.422 ± 4.369 | 69.952 ± 5.779 | 66.893 ± 7.056 |
| **PANCAN (confidence)** | 74.9 ± 5.30 | 74.2 ± 4.75 | 73.9 ± 5.00 | 72.5 ± 8.67 | 75.3 ± 6.29 |
| **TCGA** | 84.048 | 76.000 | 76.372 | 75.472 | 77.273 |

*The table summarizes classification performance of PanSubNet on the PANCAN cohort using high-confidence RNA-seq–derived labels, the entire PANCAN cohort (including lower-confidence cases), and the independent TCGA cohort. Reported metrics include area under the ROC curve (AUC), accuracy, balanced accuracy (Bal. Acc.), sensitivity, and specificity. Values for PANCAN are shown as mean ± standard deviation across cross-validation folds, while TCGA results reflect single-run external validation.*

***Table S2. Performance of AttMIL (UNI2-h backbone) for molecular subtype prediction.***

| Dataset | AUC | Accuracy | Bal. Acc. | Sensitivity | Specificity |
|---|---|---|---|---|---|
| **PANCAN (high confidence)** | 96.812 ± 1.026 | 86.131 ± 3.571 | 85.336 ± 4.718 | 77.678 ± 12.443 | 92.995 ± 6.783 |

| **TCGA** | 82.847 | 74.667 | 70.111 | 81.132 | 59.091 |
|---|---|---|---|---|---|

*This table reports the performance of the attention-based multiple instance learning (AttMIL) model with the UNI2-h feature extractor on the PANCAN high-confidence cohort and the external TCGA cohort. Metrics include AUC, accuracy, balanced accuracy, sensitivity, and specificity. PANCAN results are reported as mean ± standard deviation across cross-validation folds, whereas TCGA results correspond to external validation performance.*

***Table S3. Concordance of RNA sequencing with PanSubNet in TCGA database***

| **Metric** | **Result** | **Interpretation** |
|---|---|---|
| Total samples | 171 | Evaluation cohort |
| Overall concordance (accuracy) | 0.889 | High agreement between PanSubNet predictions and reference labels |
| Total misclassifications | 19 | Errors across both subtypes |
| Basal-like errors | 9 / 95 (9.5%) | Comparable error rate |
| Classical errors | 10 / 76 (13.2%) | No subtype-specific bias |
| Decision margin (correct predictions, median) | 0.462 | Higher model confidence |
| Decision margin (incorrect predictions, median) | 0.384 | Lower confidence |
| Statistical comparison of margins | $p$ = 0.014 (Mann–Whitney U) | Errors associated with reduced confidence |
| High-confidence errors | 6 / 19 (31.6%) | Subset of confidently misclassified cases |
| High-confidence definition | ≤ 0.10 or ≥ 0.90 | Confidence threshold |

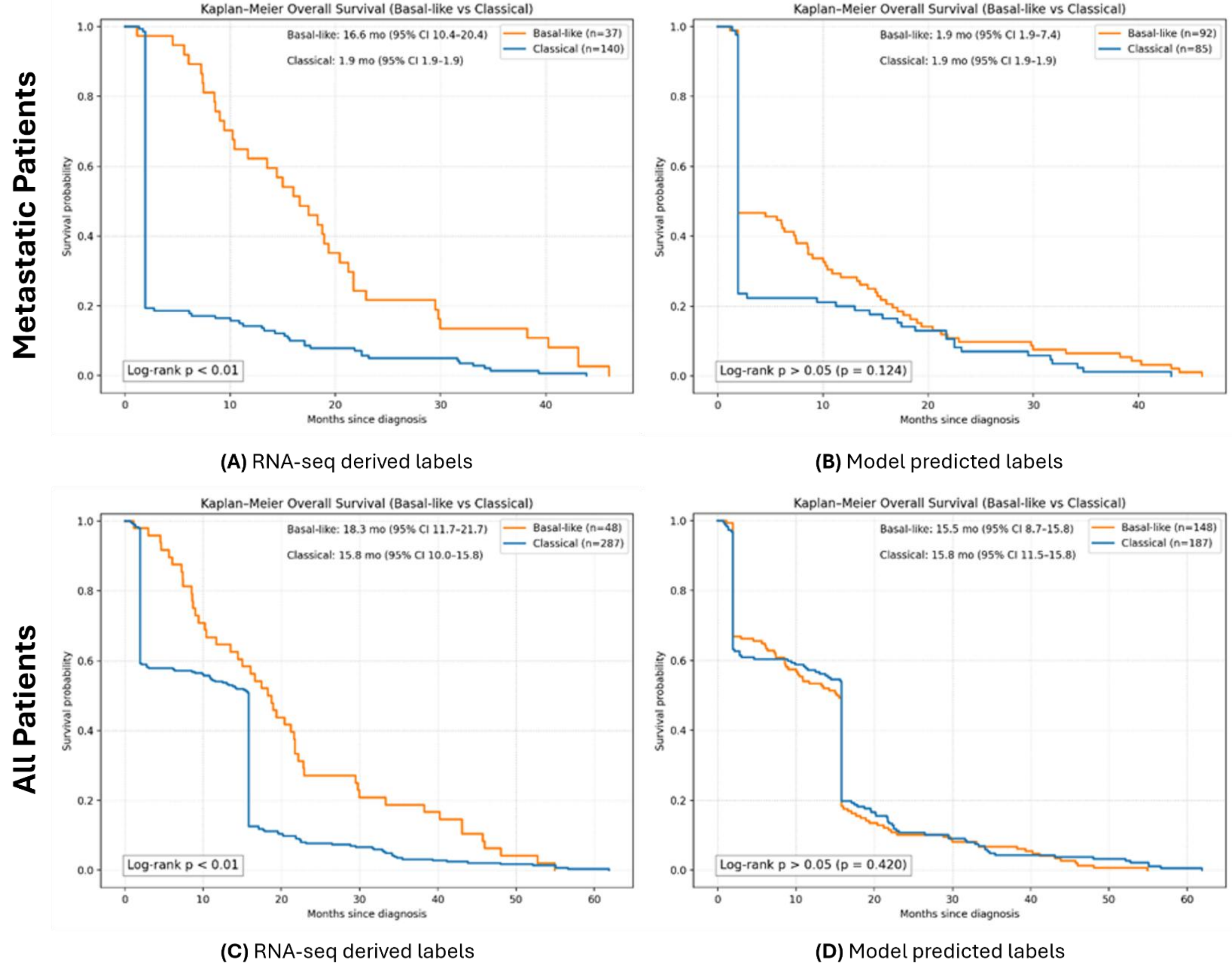

***Supplementary Figure S2. Overall survival stratified by molecular subtype in the PANCAN cohort including all cases (high- and low-confidence).*** *Kaplan–Meier curves showing overall survival (OS) stratified by Basal-like and Classical subtypes including both high- and low-confidence cases.* ***(A)*** *Metastatic patients stratified by RNA-seq–derived labels.* ***(B)*** *Metastatic patients stratified by PanSubNet-predicted labels.* ***(C)*** *Full cohort stratified by RNA-seq–derived labels.* ***(D)*** *Full cohort stratified by PanSubNet-predicted labels. Median survival times with 95% confidence intervals and log-rank test results are shown for each comparison.*

**Supplementary Methods 1: Establishing the validity of ground truth labels through transcriptomic analyses**

All transcriptomic analyses referenced in this study were performed using RNA-sequencing–derived basal-like versus classical subtype labels, and this choice was made intentionally. These RNA-seq analyses are included solely to establish that the subtype labels PanSubNet is trained to predict correspond to coherent biological programs and clinically meaningful outcomes, rather than to introduce an independent or parallel molecular analysis. PanSubNet is an H&E-based model whose sole objective is to predict the RNA-seq–defined basal-like versus classical subtype. It does not propose, redefine, or generate an independent histology-derived molecular taxonomy. Instead, the model learns histologic patterns that best approximate an established and biologically validated transcriptomic ground truth.

Accordingly, using PanSubNet-derived subtype predictions for transcriptomic characterization would not be viable. The model's predictions are explicitly optimized to reproduce RNA-seq–defined labels, and reusing those predictions to stratify gene expression would not constitute an independent biological validation. Within this framework, PanSubNet does not redefine subtype biology. Rather, it learns morphological correlates that approximate a previously established molecular phenotype. The transcriptomic analyses therefore serve as a biological anchor that justifies the prediction target itself, rather than as a discovery analysis intended to compete with the image-based model.

In this context, the image modality and the RNA-seq labels are not statistically independent, nor are they intended to be. What is independent, and what this study explicitly interrogates, is the set of histologic features learned by the model to recover the molecular subtype from routine H&E slides. The RNA-seq analyses are included to validate the biological and clinical relevance of the target label, not to supersede or parallel the image-based results. The rationale is to first establish that RNA-seq–defined basal-like and classical subtypes correspond to coherent differentiation programs, pathway activity, and prognostic behavior, and then to demonstrate that PanSubNet can accurately predict these subtypes directly from histology.

**Supplementary Methods 2: Molecular characterization reveals distinct transcriptomic programs anchored by differentiation state**

To validate the biological coherence of our ground-truth labels, we performed comprehensive molecular characterization of the classified subtypes. Principal component analysis using the top Moffitt genes demonstrated clear separation between subtypes (see Supplementary Figure S3c), with the primary component (PC1) capturing the classical-basal axis and accounting for the strongest variance. The secondary component (PC2) contributed 9% of variance, reflecting intra-subtype heterogeneity. Notably, samples clustered tightly with their respective subtype neighbors, indicating minimal transcriptomic ambiguity in high-confidence classifications.

Analysis of the most discriminative genes revealed distinct biological programs (see Supplementary Figure S3d). The classical subtype was characterized by high expression of trefoil factors (TFF1, TFF2, TFF3), which are critical for ductal differentiation and maintenance of the epithelial barrier [1]. The expression of REG4 indicated activation of PI3K/AKT and ERK/MAPK pathways, consistent with KRAS-driven oncogenesis in a differentiated cellular context [2]. Additional classical markers included SPINK1, reflecting trypsin inhibition and differentiated pancreatic function; TSPAN8 and LGALS4, involved in cellular organization and adhesion; and CEACAM5, AGR2, and CTSE, all indicators of mucin biology and functional epithelial organization [3-5].

In contrast, the basal-like cohort showed predominant expression of SCEL, an indicator of TP63 activity, and squamous differentiation with mesenchymal plasticity [6]. The expression of KLK8 and CST6 reflected loss of mucal organization and stromal infiltration, accompanied by matrix metalloproteinase activity [7, 8]. KRT6A and S100A2 signified squamous plasticity through extracellular matrix remodeling and wound repair processes [9].

Aggregate subtype program scores confirmed these patterns. Basal-like tumors showed significantly elevated basal program enrichment (see Supplementary Figure S3e), while classical tumors demonstrated higher classical program scores (see Supplementary Figure S3f). Importantly, the variance in basal marker expressions within classical tumors was greater than the variance in classical marker expressions within basal-like tumors.

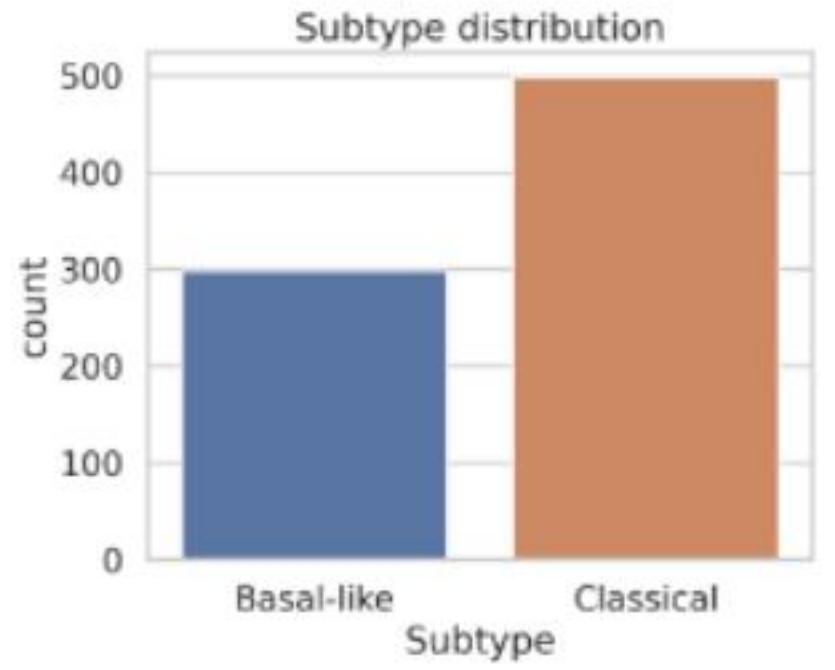

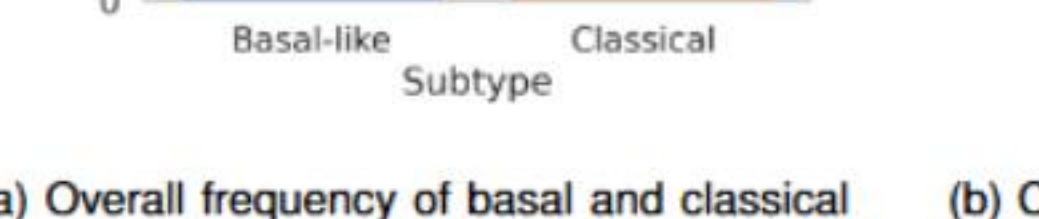

(a) Overall frequency of basal and classical subtypes across the cohort.

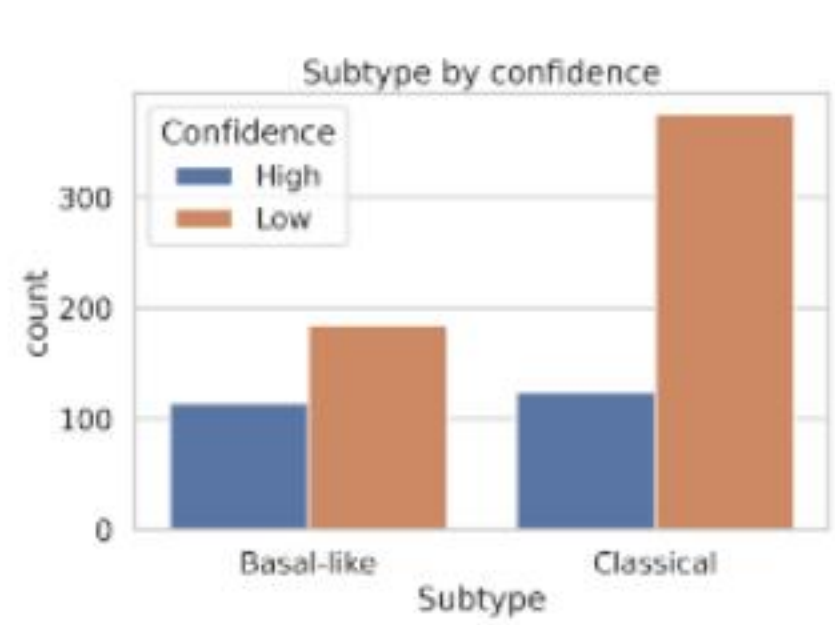

(b) Counts of samples per subtype stratified by prediction confidence.

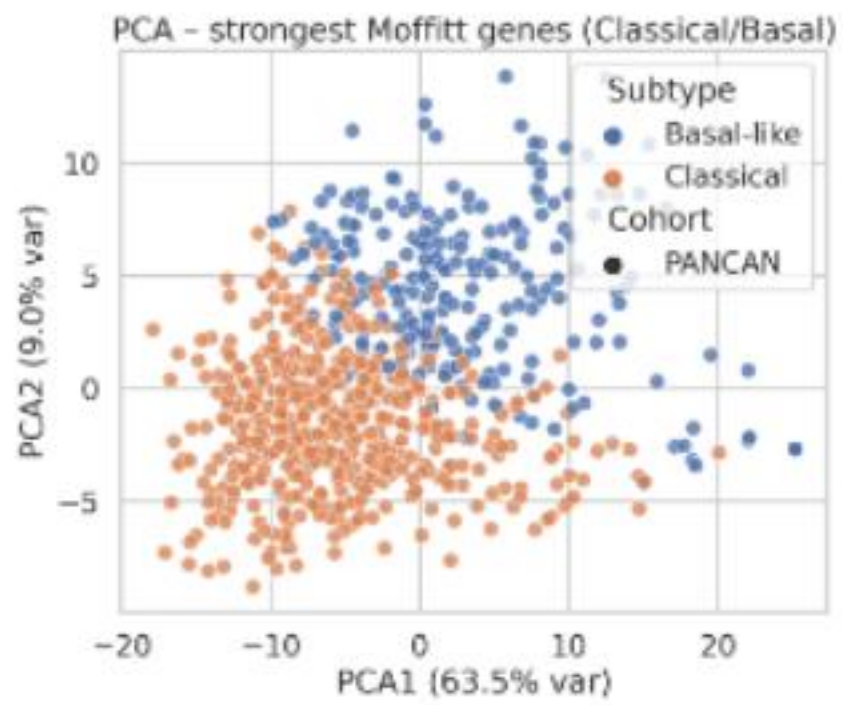

(c) PCA of samples using strong Moffitt genes, coloured by predicted subtype.

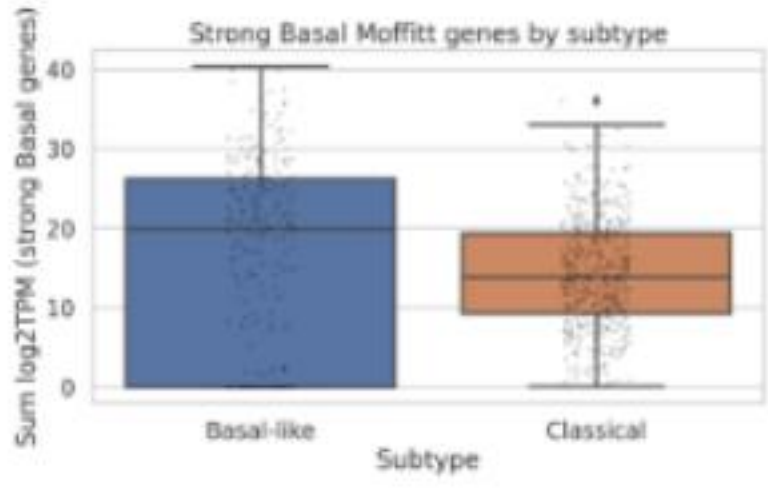

(d) Aggregate basal-program score per subtype.

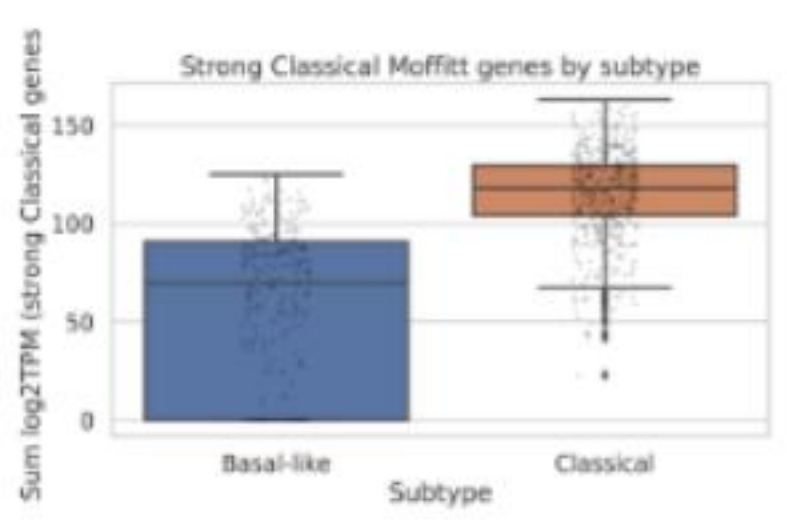

(e) Aggregate classical-program score per subtype.

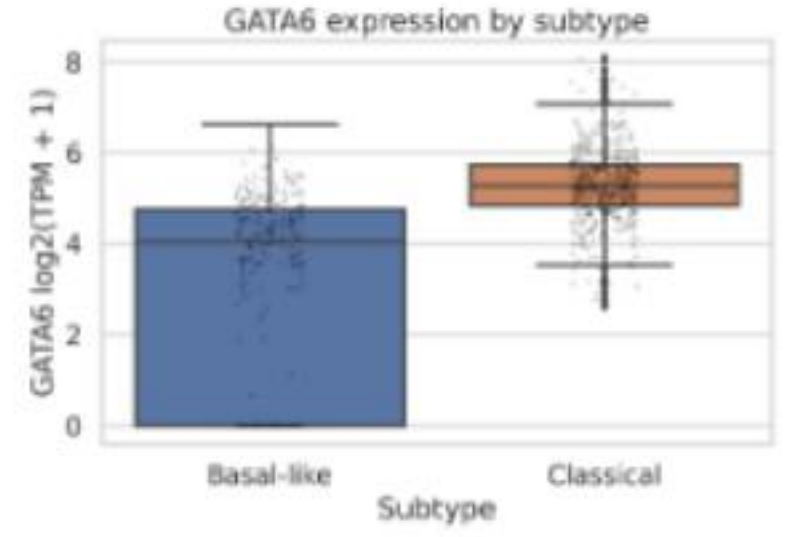

(f) GATA6 expression distribution across subtypes.

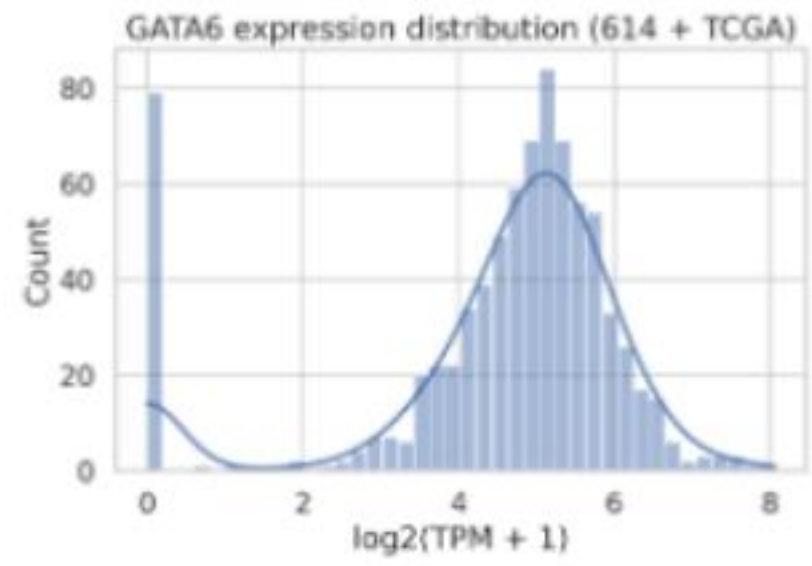

(g) Global GATA6 expression distribution.

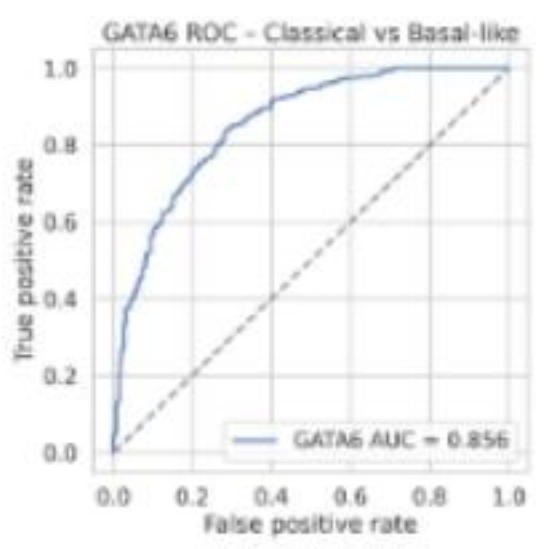

(h) ROC of GATA6 for subtype discrimination.

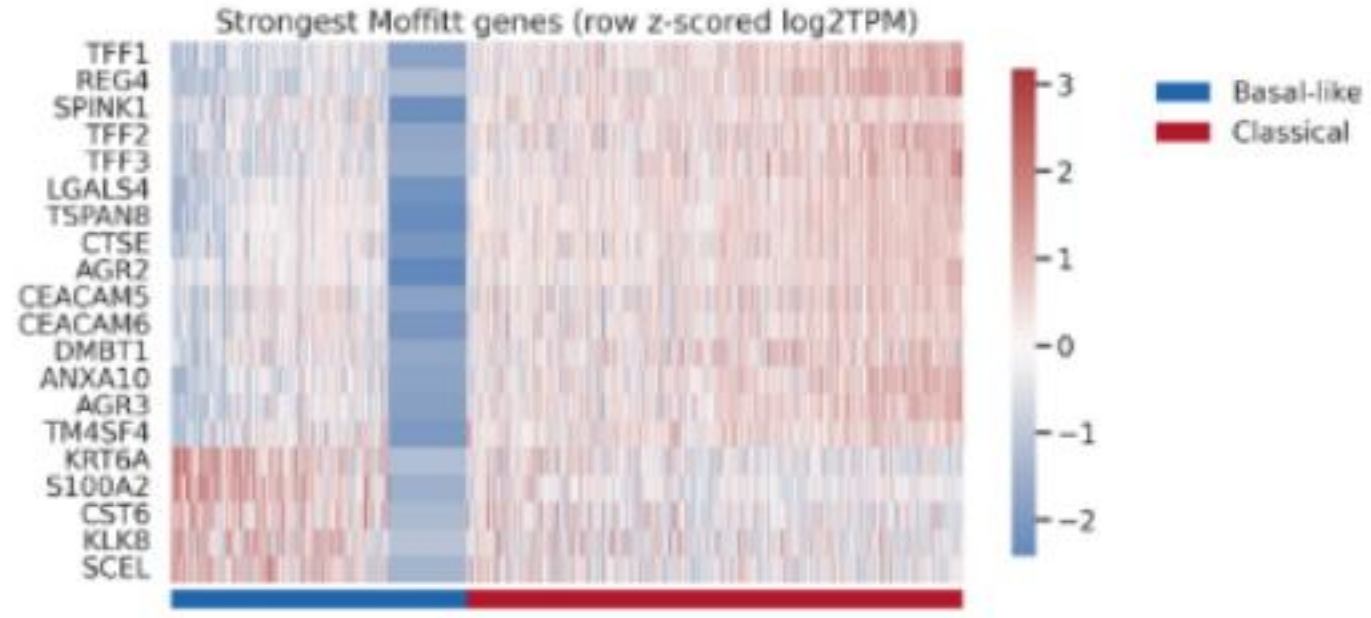

(i) Heatmap of the most subtype-discriminative Moffitt genes ordered by subtype.

***Supplementary Figure S3:*** *Integrated molecular characterization of Basal-like and Classical subtypes using PANCAN + TCGA cohorts. This figure summarizes the distribution of predicted subtypes, subtype-specific gene expression patterns, and marker-based discrimination between Basal-like and Classical pancreatic cancer.* ***(A)*** *Subtype distribution: Bar plot showing the overall frequency of Basal-like and Classical cases across the combined cohort, with Classical tumors occurring more frequently.* ***(B)*** *Subtype by prediction confidence: Stratification of predictions into high- and low-confidence groups shows that high-confidence calls are more enriched within the Classical subtype, whereas Basal-like cases include a larger proportion of low-confidence classifications.* ***(C)*** *PCA using strongest Moffitt subtype-defining genes: Principal component analysis based on the most discriminative classical/basal genes from the Moffitt signature. Points are colored by predicted subtype and shaped by cohort (PANCAN vs TCGA). Classical and Basal-like tumors show clear separation primarily along PC1. .* ***(D)*** *Basal Moffitt gene expression: Boxplots of summed log2(TPM+1) expression for strong Basal-like genes demonstrate significantly higher expression in Basal-like tumors compared to Classical tumors.* ***(E)*** *Classical Moffitt gene expression: Boxplots of summed log2(TPM+1) expression for the strong Classical genes reveal strong enrichment in Classical tumors relative to Basal-like tumors.* ***(F)*** *GATA6 expression by subtype: Boxplot of GATA6 expression showing markedly elevated expression in Classical tumors, with lower and more heterogeneous expression in Basal-like tumors.* ***(G)*** *Overall GATA6 expression distribution: Histogram and kernel density estimate illustrating the unimodal but right-skewed distribution of GATA6 expression across the combined dataset.* ***(H)*** *ROC curve for GATA6 distinguishing Classical vs Basal-like: Receiver operating characteristic analysis demonstrating strong discriminative ability of GATA6 expression alone (AUC = 0.856) for*

*separating Classical from Basal-like subtypes.* ***(I)*** *Heatmap of strongest Moffitt genes: Z-scored expression heatmap of key subtype driver genes highlights coherent Classical-high and Basal-high gene modules.*

**Supplementary Methods 3: GATA6 expression provides robust discrimination of molecular subtypes**

GATA6, a key transcription factor regulating pancreatic differentiation, showed markedly different expression patterns across subtypes (see Supplementary Figure S3g). GATA6 was used solely as a principled mechanism to clarify a small subset of transcriptionally ambiguous samples when enrichment-based Moffitt gene scores alone were insufficient to confidently assign classical or basal-like subtype. Importantly, GATA6 is not used here as a primary subtype classifier, nor is it proposed as a standalone alternative to established subtyping frameworks. Its role is limited to increasing subtype confidence within a narrowly defined ambiguous group, a use that is consistent with prior literature. In high-confidence TCGA samples, GATA6 expression was significantly higher in classical tumors than basal tumors (classical mean 4.94 vs basal mean 4.03; Mann–Whitney U $p = 2.43 \times 10^{-8}$, FDR-adjusted $p = 1.94 \times 10^{-7}$), consistent with its established role as a classical lineage marker. Classical tumors exhibited higher median GATA6 expression with lower variance, while basal-like tumors showed substantially reduced expression with greater spread. Notably, 80 samples across both cohorts showed no detectable GATA6 expression, predominantly within the basal-like group. Despite this, the majority of samples displayed moderate to high GATA6 levels.

The global distribution of GATA6 expression (see Supplementary Figure S3h) revealed a bimodal pattern consistent with its role as a lineage-defining factor. Remarkably, GATA6 expression alone achieved an AUC of 0.86 for discriminating classical from basal-like subtypes (see Supplementary Figure S3i), validating its utility as a single-gene classifier for ambiguous or intermediate-confidence cases ($p \ll .001$, DeLong's Test). Across 178 patients (93 events; basal $n = 34$, classical $n = 144$), GATA6 expression was not prognostic, with no significant difference in overall survival between high and low GATA6 groups (log-rank $p = 0.971$). In contrast, basal versus classical subtype status remained significantly associated with survival (log-rank $p = 0.0145$).This strong discriminative power supports the biological rationale for using GATA6 to refine intermediate classifications, as GATA6 activity is heavily correlated with classical phenotypes [10].

**Supplementary Methods 4: Subtype ambiguity reveals a continuous molecular spectrum with clinical implications**

To explore the continuous nature of subtype identity, we analyzed the relationship between ssGSEA-derived subtype scores ($\Delta Z$) and model prediction ambiguity. Supplementary Figure S4a shows a clear linear relationship whereby samples closer to the decision boundary ($\Delta Z \approx 0$) exhibited higher ambiguity, while those with extreme scores showed low ambiguity. When modeled as a function of absolute distance from the boundary (see Supplementary Figure S4b), ambiguity distributions were curtailed, but interestingly, peak ambiguity centered slightly toward the basal spectrum rather than at the origin.

Analysis of ambiguity versus ssGSEA subtype scores (see Supplementary Figure S4c) revealed that samples with scores near 0.5 enrichment displayed the highest ambiguity, indicating genuine intermediate states. Importantly, no high-confidence samples ($\Delta Z > 1$) demonstrated elevated ambiguity, confirming an inverse relationship between subtype confidence and prediction uncertainty.

GATA6 expression aligned strongly with the subtype axis (see Supplementary Figure S4d), with classical samples showing elevated GATA6 and basal-like samples showing reduced or absent expression. However, GATA6 expression versus ambiguity (see Supplementary Figure S4e) showed substantial overlap in the 0.6–0.8 ambiguity range across varying GATA6 levels.

Principal component analysis of samples using strong Moffitt genes, colored by ambiguity (see Supplementary Figure S4f), demonstrated that high-ambiguity samples clustered near the interface between classical and basal regions, while low-ambiguity samples formed tight, distinct clusters.

When analyzing variance in the top 10 Moffitt genes per subtype, higher variance corresponded to lower ambiguity, though most samples clustered within 5 variance units of the origin. High-variance samples at either tail were enriched for low-ambiguity cases, indicating that the subtype split is driven by variability in gene expression rather than consistent uniform enrichment.

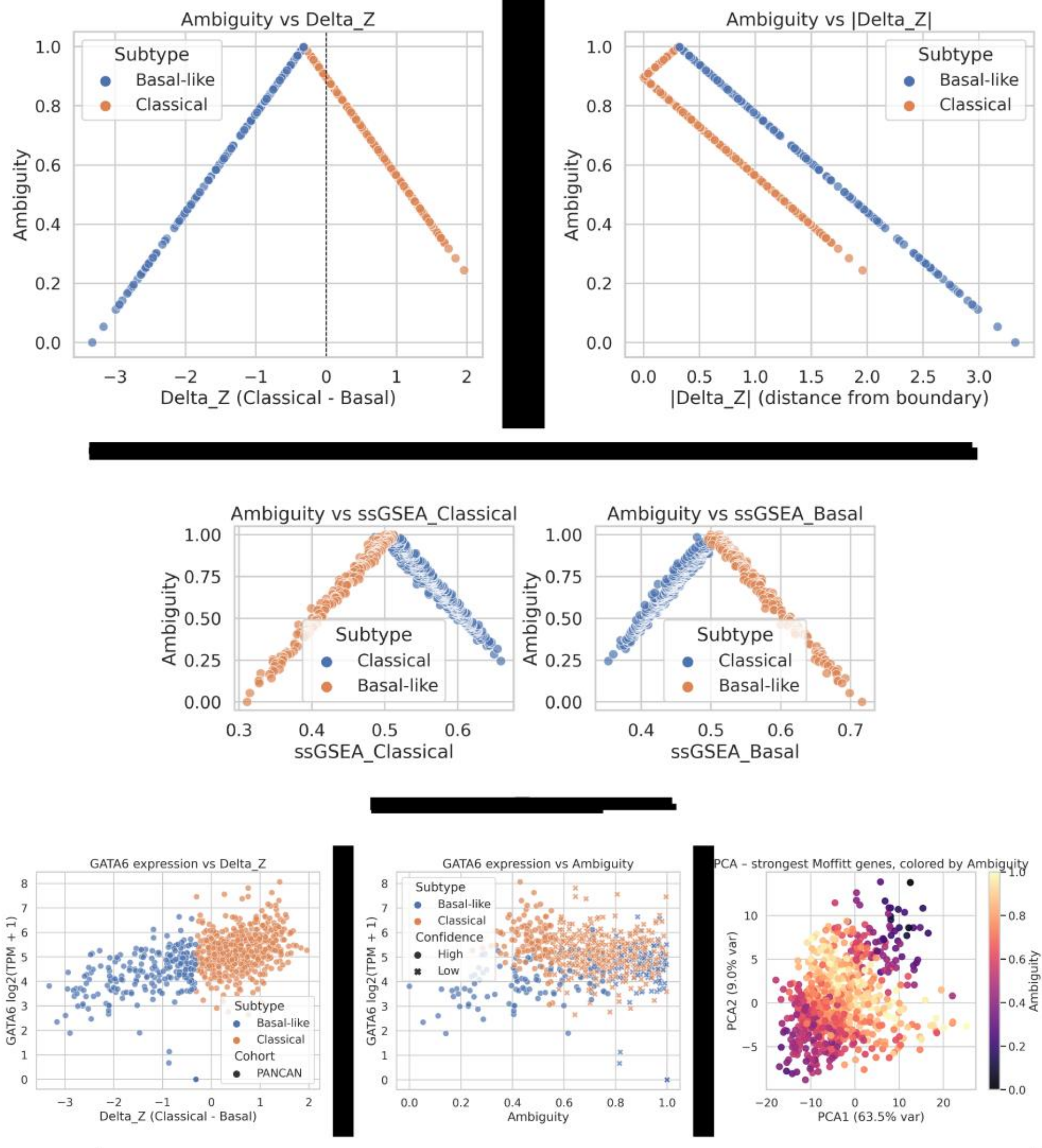

***Supplementary Figure S4:*** *Ambiguity score behavior, subtype separation, and biological correlates across PANCAN + TCGA samples. This figure characterizes the ambiguity score and its relationship to subtype-defining molecular features, ssGSEA signatures, GATA6 expression, and PCA structure.* ***(A)*** *Ambiguity vs. ΔZ (Classical – Basal score). Ambiguity is maximal near ΔZ = 0 (the subtype decision boundary) and decreases linearly as tumors diverge toward strongly Classical (positive ΔZ) or strongly Basal-like (negative ΔZ). Basal-like tumors populate negative ΔZ values, whereas Classical tumors populate the positive side.* ***(B)*** *Ambiguity vs. |ΔZ| (absolute distance from boundary). Absolute distance from the decision boundary is inversely correlated with ambiguity. Tumors farther from the boundary show exceptionally low ambiguity, whereas tumors near the center show high uncertainty.* ***(C)*** *Ambiguity vs. ssGSEA Classical enrichment. Classical tumors demonstrate higher ssGSEA enrichment for the Classical signature and correspondingly lower ambiguity. Basal-like tumors have reduced Classical ssGSEA scores and higher ambiguity near the transition zone.* ***(D)*** *Ambiguity vs. ssGSEA Basal enrichment. Basal-like tumors exhibit higher ssGSEA enrichment for the Basal signature, again with lower ambiguity as they move deeper into the Basal-like state. Classical tumors with weak Classical signatures show elevated ambiguity. Together, (C–D) validate that the ambiguity score aligns with biologically meaningful continuous variation, rather than classification noise.* ***(E)*** *GATA6 expression vs. ΔZ. GATA6 expression increases sharply with ΔZ, showing elevated levels in Classical tumors and low levels in Basal-like tumors. The gradient further supports GATA6 as a continuous indicator of subtype state.* ***(F)*** *GATA6 expression vs. ambiguity. Low-ambiguity tumors show clear subtype extremes: high GATA6 for Classical and low GATA6 for Basal-like. High-ambiguity tumors cluster near intermediate GATA6 expressions and include many low-confidence predictions.* ***(G)*** *PCA of strongest Moffitt genes, colored by ambiguity. Samples with high ambiguity cluster near the interface between Classical and Basal regions in PCA space. Clear subtype extremes exhibit low ambiguity and occupy opposite ends of the major expression gradient.*

**Supplementary Methods 5: DNA damage repair gene expression links molecular subtype to chemosensitivity**

Given the established association between DDR deficiency and platinum sensitivity, we examined the expression of key DDR genes across subtypes [11]. The Gene expressions of BRCA1/2, RAD51, ATM, PALB2, and CHEK1 were measured across all patients classified into the basal-like and classical category. Gene expression across all genes was summated and Z-scored across both subtypes and all cohorts into a stratified DDR expression score to visualize trends in gene expression within the basal and classical cohorts that indicate chemosensitivity. Median BRCA1 expression was similar between classical and basal-like cohorts (see Supplementary Figure S5a), but the classical cohort showed substantially greater spread, with the 25th percentile near zero expression. In contrast, basal-like tumors exhibited a tighter distribution centered at moderate expression. BRCA2 expression (see Supplementary Figure S5b) mirrored this pattern, though classical tumors displayed even lower expression at the 25th percentile.

Analysis of PALB2 (see Supplementary Figure S5c), RAD51 (see Supplementary Figure S5d), ATM (see Supplementary Figure S5e), and CHEK1 (see Supplementary Figure S5f) revealed similar trends. Classical tumors generally showed more stratified expression with greater variance, while basal-like tumors exhibited tighter distributions. Notably, median ATM expression was marginally higher in classical tumors, whereas median CHEK1 expression was lower in classical compared to basal-like tumors. The composite DDR score (see Supplementary Figure S5g), computed by aggregating expression across all DDR genes, showed that classical tumors had more stable, clustered expression profiles, whereas basal-like tumors displayed greater variability with numerous low-expression outliers.

Heatmap analysis ordered by $\Delta Z$ (see Supplementary Figure S5h) revealed that a large subset of samples expressed minimal or no DDR genes, distributed across both subtypes. When DDR score was plotted against $\Delta Z$ (see Supplementary Figure S5i), a linearly negative relationship emerged, though scattered with outliers predominantly in the basal-like cohort. This indicates that higher classical subtype confidence correlates with lower DDR expression on average. In high-confidence TCGA samples, multiple DNA damage response (DDR) genes exhibited significant subtype-specific expression differences independent of GATA6. CHEK1, RAD51, and BRCA2 showed higher expression in basal tumors compared with classical tumors (CHEK1: Welch $p = 1.35 \times 10^{-4}$, $FDR = 5.41 \times 10^{-4}$; RAD51: Mann–Whitney U $p = 8.28 \times 10^{-4}$, $FDR = 2.21 \times 10^{-3}$; BRCA2: Mann–Whitney U $p = 1.80 \times 10^{-2}$, $FDR = 3.59 \times 10^{-2}$), whereas BRCA1, PALB2, and ATM were not significantly different after correction.

Modeling DDR score against prediction ambiguity (see Supplementary Figure S5j) showed no clear separation, though a substantial proportion of highly ambiguous samples clustered near zero DDR score, indicating no variability in DDR gene enrichment. A composite DDR score demonstrated a modest but significant shift between basal and classical tumors (Mann–Whitney U $p = 3.29 \times 10^{-2}$; $FDR = 5.26 \times 10^{-2}$). In survival analysis of 178 patients (93 events), high versus low DDR composite scores significantly stratified overall survival (log-rank $p = 0.0031$), in contrast to GATA6 expression, which was not prognostic, indicating that DDR activity captures prognostic information not explained by lineage alone.

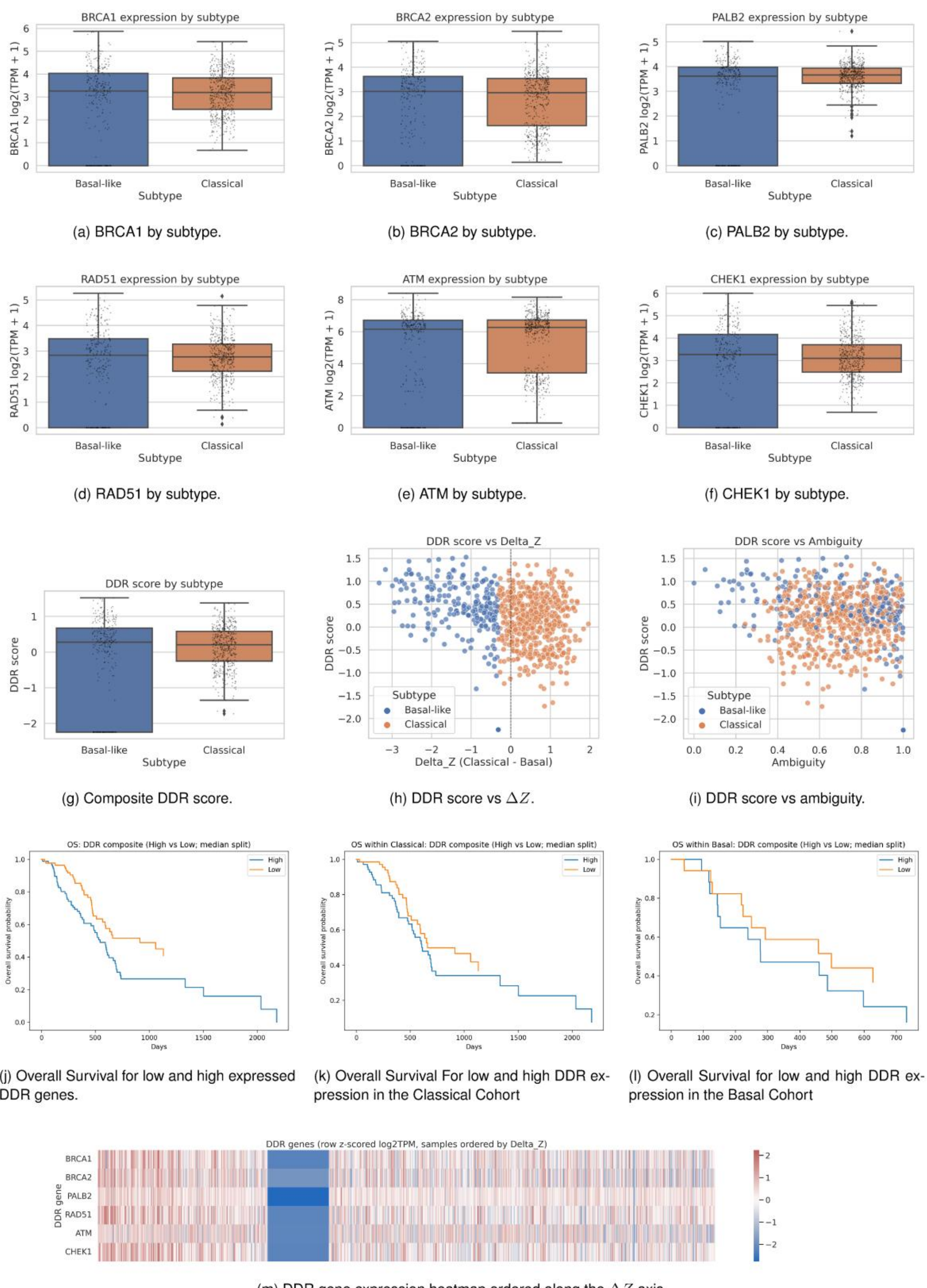

(a) BRCA1 by subtype.

(b) BRCA2 by subtype.

(c) PALB2 by subtype.

(d) RAD51 by subtype.

(e) ATM by subtype.

(f) CHEK1 by subtype.

(g) Composite DDR score.

(h) DDR score vs $\Delta Z$.

(i) DDR score vs ambiguity.

(j) Overall Survival for low and high expressed DDR genes.

(k) Overall Survival For low and high DDR expression in the Classical Cohort

(l) Overall Survival for low and high DDR expression in the Basal Cohort

(m) DDR gene expression heatmap ordered along the $\Delta Z$ axis.

***Supplementary Figure S5:*** *DNA Damage Response (DDR) gene expression patterns across Basal-like and Classical tumors and their relationship to subtype continuum metrics. This figure examines expression of key homologous recombination and DDR genes across subtypes, integrates these genes into a composite DDR score, and evaluates how DDR activity varies across the Classical–Basal continuum. (**Top two rows:** Individual DDR gene expression by subtype) Boxplots display log2(TPM+1) expression for canonical homologous recombination repair (HRR) and DDR regulators across Basal-like and Classical tumors: **(A)** BRCA1, **(B)** BRCA2, **(C)** PALB2, **(D)** RAD51, **(E)** ATM, **(F)** CHEK1 Across all six genes, Classical tumors consistently show*

*higher expression than Basal-like tumors. Basal-like tumors exhibit lower overall DDR component expression with greater heterogeneity, consistent with reduced HRR pathway activity. (**Third row:** Composite DDR score distributions and continuous relationships)* ***(G)*** *DDR score by subtype. A composite DDR score (computed as the mean z-scored expression of BRCA1, BRCA2, PALB2, RAD51, ATM, and CHEK1) is significantly lower in Basal-like tumors and higher in Classical tumors, mirroring individual gene-level patterns.* ***(H)*** *DDR score vs. ΔZ. DDR activity increases monotonically with ΔZ (Classical – Basal). Classical tumors (positive ΔZ) show clear enrichment for DDR signaling, whereas Basal-like tumors (negative ΔZ) cluster toward reduced DDR activity.* ***(I)*** *DDR score vs. Ambiguity. Higher ambiguity is associated with lower DDR activity, particularly among Basal-like tumors. Classical tumors with strong DDR signaling display low ambiguity, further supporting the biological interpretability of the ambiguity metric.* ***(J–L)*** *Kaplan–Meier overall survival analyses stratified by high versus low DNA damage response (DDR) composite score (median split), shown for the full cohort (j), the classical subtype (k), and the basal subtype (l). Survival associations differ by subtype, with significant stratification observed in the full cohort and classical tumors, but not in basal tumors.* ***(M)*** *Heatmap of DDR genes (row-z-scored log2TPM), samples ordered by ΔZ. Samples are arrayed from strongly Basal-like (left) to strongly Classical (right). DDR gene expression shows a coordinated gradient: Basal-like tumors exhibit systematic downregulation (blue), while Classical tumors demonstrate cohesive upregulation (red). This provides visual confirmation of DDR pathway strengthening along the Classical direction.*

**Supplementary Methods 6: Gene ontology and pathway analysis reveal distinct functional programs**

To assess whether the model captured biologically functional rather than purely statistical features, we examined the biological programs represented by subtype-associated genes using Gene Ontology (GO) and KEGG pathway enrichment analyses of the top 25 classical and basal-like genes [12, 13]. GO Biological Process enrichment (see Supplementary Figure S6a, c–d) showed that classical-associated genes were predominantly linked to digestive system processes, digestion, and epithelial tissue development. These enrichments were driven by genes such as REG4, TFF1/2/3, CEACAM5, and LGALS4, which are established mediators of mucin secretion, epithelial barrier integrity, and pancreatic ductal function [54]. Additional enrichment was observed for epithelial structural maintenance and gastrointestinal tract development, with weaker associations involving metabolic regulation and anoikis resistance.

In contrast, basal-like–associated genes exhibited strong enrichment for tissue development, epidermal differentiation, and wound-associated programs. Notably, a substantial fraction of genes annotated to tissue development also mapped to wound healing, keratinization, and extracellular matrix remodeling, consistent with the regenerative, mesenchymal, and stress-responsive phenotypes characteristic of basal-like tumors [55].

Consistent with these findings, KEGG pathway analysis (see Supplementary Figure S6b, e–f) identified significant enrichment of cornified envelope formation within the basal-like gene set, involving eight highly significant genes, a pathway closely associated with squamous differentiation and wound repair [55]. In comparison, only a single classical-associated gene showed weak enrichment for this pathway. These molecular programs closely parallel the squamous histomorphology observed in basal-like pancreatic ductal adenocarcinoma [14], demonstrating that the model recovers subtype-specific biological processes rather than relying on uninterpretable image correlations.

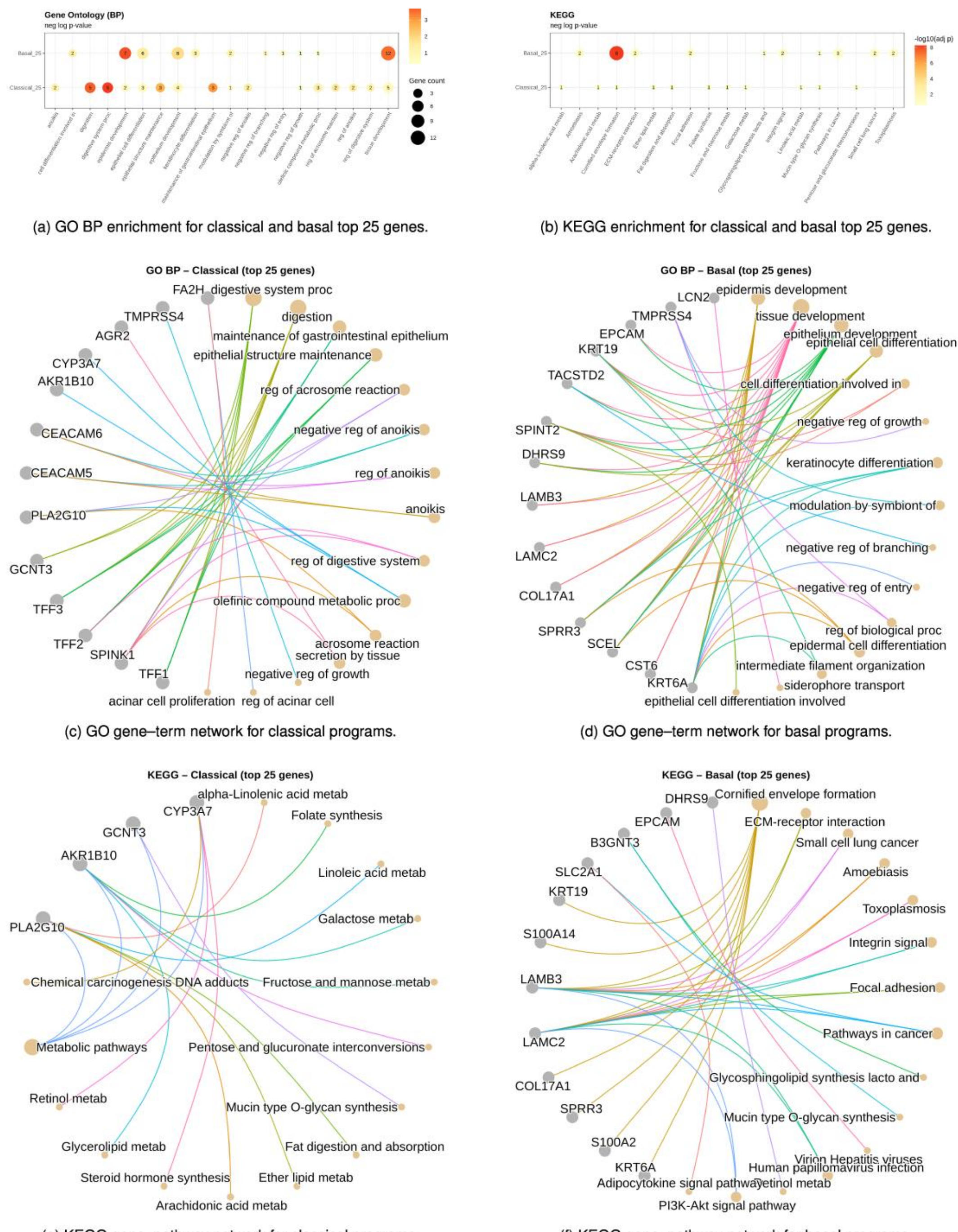

(a) GO BP enrichment for classical and basal top 25 genes.

(b) KEGG enrichment for classical and basal top 25 genes.

(c) GO gene–term network for classical programs.

(d) GO gene–term network for basal programs.

(e) KEGG gene–pathway network for classical programs.

(f) KEGG gene–pathway network for basal programs.

***Supplementary Figure S6:*** *Functional enrichment analysis of Classical and Basal-like subtype–specific genes reveal distinct biological programs and pathway specializations. This figure summarizes Gene Ontology (GO) and KEGG enrichment patterns for the top differentially expressed genes defining the Classical and Basal-like pancreatic cancer subtypes. Bubble plots provide high-level enriched terms, while chord diagrams map individual genes to their associated pathways, highlighting subtype-specific biological roles.* ***(A)*** *Gene Ontology Biological Process (GO BP) enrichment. Classical subtype genes are enriched for digestive, metabolic, acinar-cell, and epithelial maintenance processes. Basal-like genes show enrichment in keratinocyte differentiation, epithelial development, epidermal structure, and stress- or injury-related processes. Bubble color represents –log10 adjusted p-value; size represents gene count.* ***(B)*** *KEGG pathway enrichment. Classical subtype genes preferentially map to lipid metabolism, xenobiotic metabolism, steroid synthesis, and other homeostatic epithelial functions. Basal-like subtype genes show strong enrichment in ECM–receptor interaction, focal adhesion, PI3K–Akt signaling, pathways in cancer, and infection-related KEGG pathways, consistent with a more mesenchymal and stress-associated transcriptional state.* ***(C)*** *Classical subtype – top 25 GO BP genes. Chord diagram links*

*Classical-specific genes to enrich biological processes such as digestive system function, epithelial structure maintenance, acinar cell regulation, and olefinic compound metabolism. Prominent genes include CEACAM6, TMPRSS4, AGR2, CYP3A7, and AKR1B10, emphasizing classical epithelial identity and metabolic specialization.* ***(D)*** *Basal-like subtype – top 25 GO BP genes. Basal-like genes connect to differentiation programs (epidermal, keratinocyte, epithelial), negative regulation of growth, branching morphogenesis, and modulation of biological processes. Key drivers include KRT6A, KRT19, LCN2, LAMC2, and SPRR3, reflecting a wound-like, regenerative, and injury-responsive phenotype.* ***(E)*** *Classical subtype – top 25 KEGG genes. Classical genes map to lipid metabolism (arachidonic acid, alpha-linolenic acid), carbohydrate metabolism (galactose, fructose/mannose), retinol metabolism, steroid hormone synthesis, and chemical carcinogenesis pathways. These enrichments match a differentiated, metabolically active epithelial lineage.* ***(F)*** *Basal-like subtype – top 25 KEGG genes. Basal-like genes connect to ECM–receptor interaction, focal adhesion, integrin signaling, PI3K–Akt signaling, pathways in cancer, and viral/infection-related pathways. Genes such as LAMC2, COL17A1, DHRS9, S100A2, and EPCAM highlight extracellular matrix remodeling, motility, and stress-response biology.*